\documentclass[12pt]{article}
\usepackage{amsmath, amssymb,bm,amsthm}
\usepackage{geometry}

\usepackage{algorithm}
\usepackage{algpseudocode}

\usepackage{hyperref}
\usepackage{url}
\usepackage{comment}
\usepackage{graphicx}
\usepackage{caption}
\usepackage{subfigure}
\usepackage{mathtools}

\usepackage[dvipsnames]{xcolor}
\newcommand{\misha}[1]{{\color{PineGreen}\underline{\bf Misha says:} #1}}

\usepackage[most]{tcolorbox}
\usepackage{booktabs}

\usepackage{pgfplots,tikz}

\numberwithin{equation}{section}

\usepackage[backend=biber, style=ieee, sorting=none]{biblatex}
\newcommand{\tar}{\text{\tiny tar}}

\title{Generative Stochastic Optimal Transport:\\ Guided Harmonic Path-Integral Diffusion}

\author{Michael (Misha) Chertkov\\   Applied Mathematics Graduate Interdisciplinary Program\\ and Department of Mathematics, University of Arizona}
\date{\today}

\begin{document}

\maketitle

\begin{abstract}

We introduce \emph{Guided Harmonic Path–Integral Diffusion} (GH--PID), a linearly– solvable framework for \emph{guided} stochastic optimal transport (SOT) with a \emph{hard} terminal distribution and \emph{soft}, application–driven path costs.  A low–dimensional guidance protocol $\Gamma_t=\{\beta_t,\nu_t\}$ shapes the trajectory ensemble while preserving analytic structure: the forward and backward Kolmogorov equations remain linear, the optimal score admits an explicit Green–function ratio, and Gaussian–Mixture Model (GMM) terminal laws yield closed–form expressions for the backward probe $\hat y(t;x)$ and the optimal drift $u_t^\ast(x)$.  This enables stable sampling and differentiable protocol learning under exact terminal matching.

We develop guidance–centric diagnostics -- path cost, centerline adherence, variance flow, and drift effort -- that make GH-PID an interpretable variational ansatz for empirical SOT. Three navigation scenarios illustrated in 2D: (i) Case A: hand-crafted protocols revealing how geometry and stiffness shape lag, curvature effects, and mode evolution; (ii) Case B: single-task protocol learning, where a PWC centerline is optimized to minimize integrated cost; (iii) Case C: multi-expert fusion, in which a commander reconciles competing expert/teacher trajectories and terminal beliefs through an exact product-of-experts law and learns a consensus protocol. Across all settings, GH-PID generates geometry-aware, trust-aware trajectories that satisfy the prescribed terminal distribution while systematically reducing integrated cost.


\end{abstract}

\tableofcontents

\section{Introduction}\label{sec:intro}
\subsection{Diffusion with paths in mind} 

Diffusion samplers have become a standard route for drawing samples from complex, continuous distributions~\cite{ho_denoising_2020,song_score-based_2021}. While most practical pipelines are score–based diffusions (SBDs) that learn the \emph{marginal} score $\nabla_x \log p_t(x)$, a complementary \emph{bridge} viewpoint casts sampling as a finite–time, non-autonomous transport from a tractable reference to the target~\cite{peluchetti_non-denoising_2021,behjoo_space-time_2024}. This “bridge–diffusion’’ perspective connects directly to Schrödinger bridges~\cite{schrodinger_sur_1932}, Feynman–Kac path integrals~\cite{feynman_space-time_1948,kac_distributions_1949}, and non-equilibrium work relations~\cite{jarzynski_equilibrium_1997,chernyak_dynamical_2005}.

\subsection{The Principle Problem in Focus} 

We consider a controlled Itô \emph{Stochastic Ordinary Differential Equation} (SODE) over a finite time horizon:
\begin{equation}
\label{eq:sde}
d x_t \,=\, u_t(x_t)\,dt \;+\; dW_t,\qquad x_0 = 0 \in \mathbb{R}^d,\quad t\in[0,1],
\end{equation}
where $\{W_t\}$ is a standard $d$-dimensional Brownian motion and $u_t:\mathbb{R}^d\to\mathbb{R}^d$ is the (to-be-determined) drift/score field. We adopt the aforementioned path-centric lens of the bridge-diffusion: the terminal law for the probability density, $p_{t=1}(x_1)$ is strictly enforced (hard constraint) by the target density
\begin{equation}
\label{eq:target}
p_{t=1}(x_1)=p^{(\tar)}(x_1) \;\propto\; \exp\!\big(-E(x_1)\big),\qquad x_1\in\mathbb{R}^d,
\end{equation}
where $E:\mathbb{R}^d\to\mathbb{R}$ is a given energy (known up to an additive normalization constant). But the entire \emph{trajectory} -- $x_{0\to 1}$ governed by Eq.~(\ref{eq:sde}) -- is also central (soft constraint). Overall we address the following principle {\bf Stochastic Optimal Transport (SOT)} problem:
\begin{equation}
\min_{u_{0\to 1}} J(u_{0\to 1}),\quad J(u_{0\to 1})\doteq \int_0^1 \mathbb{E}[\,C_t(x_t;u_t(x_t))\mid \text{Eqs.~(\ref{eq:sde},\ref{eq:target})}\,]\,dt,
\label{eq:principle-SOT}
\end{equation}
where $C_t(x;u)$ is a cost-to-go function which encodes application-specific soft constraints over the path, such as trajectory smoothness, dynamic safety in the case of navigation or cost of transformation in the case of paths associated with chemical transformations. 

Solving principle SOT for a generic cost-to-go is challenging -- in fact, hardly possible exactly. Instead, in Section \ref{subsec:caseB} we develop an approximate {\bf variational} approach (for an exemplary cost to go which depends only on $x_t$, and not on $u_t(x_t)$) -- based on the {\bf Path Integral Diffusion}  \cite{behjoo_harmonic_2025} -- and illustrate its utility on applications in spatio-temporal navigation in 2D.

\subsection{PID as linearly-solvable SOC and dual SOT} 

The PID framework of \cite{behjoo_harmonic_2025} sits at the intersection of stochastic control and transport. On the control side, PID leverages \emph{linearly solvable} stochastic control~\cite{mitter_non-linear_1981,kappen_path_2005,e_todorov_linearly-solvable_2007,dvijotham_unifying_2012,chernyak_stochastic_2014,kappen_adaptive_2016} to express optimal drifts via ratios of forward/backward Green functions of linear Kolmogorov–Fokker–Planck (KFP) operators. On the SOT side, PID was introduced as a particular {\bf "integrable"} sub-family of the principle SOT family defined by Eq.~(\ref{eq:principle-SOT}). Specifically in a special (zero drift and gauge) case of PID  the Cost-to-Go in the principle formulation of Eq.~(\ref{eq:principle-SOT}) is decomposed into kinetic and potential parts:
\begin{equation}\label{eq:C-PID}
    C^{(PID)}_t(x;u)=\frac{u^2}{2}+V_t(x).
\end{equation}
Notably an even earlier version of the PID -- with zero potential $V_t=0$ -- and one aligned with the control-as-inference and SB/SDE bridges view of Generative AI was introduced in \cite{tzen_theoretical_2019} (see also references therein for discussion of the topic's pre-history). In addition to discussing general integrable structures -- with the degrees of freedom associated with arbitrary drifts, gauges and potential within respective SOT formulations, \cite{behjoo_harmonic_2025} focused on the simplest, but already non-trivial case, with zero drift and gauge and quadratic/harmonic isotropic potential -- $V_t(x)=\beta_t x^2/2$, with a time-independent stiffness coefficient $\beta_t$. Then in \cite{chertkov_adaptive_2025} a more general case of time-dependent stiffness coefficient, $\beta_t$ was analyzed. In this paper -- and on the way to an approximate (variational) solution of the principle challenge (\ref{eq:principle-SOT}) -- we also made the next step in the development of the PID methodology --- extend the class of integrable potentials and consider {\bf Guided Harmonic (GH) PID}, where we allow the harmonic potential not only to change stiffness $\beta_t$ in time but also location of the minimum $\nu_t$.

\subsection{Guided Harmonic PID: protocol, potential, and scope} 

We extend harmonic PID (H-PID) to a \emph{Guided} setting by introducing the \emph{Guided Harmonic} (GH) potential
\begin{equation}
\label{eq:guided_potential}
V_t(x) \;=\; \frac{\beta_t}{2}\,\big\|x-\nu_t\big\|^2,
\end{equation}
where $\beta_t>0$ is a \emph{stiffness}  and $\nu_t\in\mathbb{R}^d$ is a \emph{guide} (moving center), which we combine into the collective protocol
\begin{equation}
\label{eq:Gamma_def}
\Gamma_t \;\doteq\; (\beta_t,\nu_t), 
\qquad 
\Gamma_{0\to 1} \;\doteq\; (\Gamma_t: t\in[0,1]),
\end{equation}
steering  the flow in state space -- while preserving linear solvability. The result is a mixed SOT program with a \emph{hard} terminal constraint (match the target) and \emph{soft} path costs (encode guidance).


\subsection{Design principles of GH-PID and its analytic levers} 

In PID the terminal law $p^{(\mathrm{tar})}$ and the path costs are, in principle, arbitrary. We specialize both objects to preserve linear solvability and transparency of the optimal drift. For the terminal law we adopt a Gaussian–Mixture Model (GMM) (as in~\cite{chertkov_adaptive_2025}), which yields closed-form posteriors. For the path costs we use the guided quadratic potential (\ref{eq:guided_potential}) with Piece-Wise-Constant (PWC) schedules for $\Gamma_t$ -- this gives closed-form interface “kicks’’ and robust time stepping. In combination, GMM targets and guided PWC quadratic potential allow us to compute the optimal control explicitly according to the compact GH–PID identity
\begin{equation}
\label{eq:ustar_compact_forwardref}
u_t^\ast(x;\Gamma) = b_t^{(-)}\Bigl(\hat y(t;x;\Gamma)-\mu_t(x;\Gamma)\Bigr),
\end{equation}
interpreted as a re-weighted mismatch between two time and state dependent expectations of the target state -- with analytic expressions for all the functions involved -- see Appendix \ref{sec:Riccati} for the Green (functions)/Riccati (equations) machinery and Appendices \ref{sec:yhat} and \ref{sec:ustar} for the explicit/analytic formulas for maps from current state to predicted state and to velocity (optimal control). Practically, $\beta_t$ and $\nu_t$ are interpretable \emph{levers} for protocol design, and the analyticity enables principled trade-offs among terminal fidelity, numerical conditioning (at the inference time), and path semantics.

\subsection{Contributions of this paper}

\begin{itemize}
\item \textbf{Guided, linearly–solvable mixed SOT.} We formulate a \emph{guided} Stochastic Optimal Transport (SOT) problem with a \emph{hard} terminal distribution and \emph{soft}, application–driven path costs. The guidance protocol $\Gamma_t=\{\beta_t,\nu_t\}$ modulates stiffness and centerline preferences. We develop the full analytic GH–PID toolkit:  (i) optimal score expressed as ratios of linear KFP Green functions;  (ii) closed–form Riccati and Green–function updates for time–varying $\beta_t$ and moving $\nu_t$;  (iii) explicit PWC schedules with interface kicks;  (iv) a compact drift representation (Eq.~\ref{eq:ustar_compact_forwardref})  and analytic $\hat y(t;x)$ fo Gaussian–mixture (GMM) terminal laws.  These ingredients form a linearly–solvable backbone for guided diffusion under hard terminal constraints. See Section~\ref{sec:formulation} and Appendices~\ref{sec:Riccati}–\ref{sec:ustar}.

\item \textbf{Empirical SOT via protocol optimization with diagnostics.} We show that GH–PID serves as a structured {\bf variational} ansatz for the principal SOT objective \eqref{eq:principle-SOT}. For any cost-to-go function $C_t(x,u)$ we minimize the time–integrated functional \eqref{eq:principle-SOT} under exact terminal matching. This yields \emph{protocol learning} over $\Gamma_t$ using stable, interpretable guidance–centric diagnostics: path cost, variability, adherence, control effort, and sensitivity. See Section~\ref{sec:diagnostics}. We choose to illustrate the approach -- in Section~\ref{sec:navigation} -- on an exemplary 2D navigation case where the cost-to-go depends only on $x$ (and not on $u$).

\item \textbf{Three navigation examples of increasing sophistication.} We specialize GH–PID -- as a variational ansatz -- to 2D navigation, where hard terminal constraints and soft trajectory shaping are both natural and essential. Section~\ref{sec:navigation} and Appendix~\ref{app:navigation} develop: (i) \emph{Case A}: fully hand–crafted protocols for funnel and tunnel geometries, demonstrating analytic path shaping;   (ii) \emph{Case B}: protocol optimization for a fixed task-driven GMM terminal law (multi-exit navigation);   (iii) \emph{Case C}: multi–task navigation via a product–of–experts terminal law and joint protocol learning. These examples illustrate GH–PID as a flexible, interpretable mechanism for mixed SOT that captures corridor adherence, risk avoidance, and multi-objective navigation.

\item \textbf{Numerical case studies.} Section~\ref{sec:cases} reports aggregated quantitative results for all three navigation settings, highlighting reduction of integrated path cost at fixed terminal fidelity, improved safety and adherence, and the ability to tune path properties through the low-dimensional protocol $\Gamma$.
\end{itemize}

\section{Guided Harmonic PID: Problem Formulation}
\label{sec:formulation}
While the central goal of this work is to solve the principal SOT problem \eqref{eq:principle-SOT}, our approach is \emph{not} to solve it directly. Instead, we seek a linearly-solvable Guided Harmonic PID (GH-PID) model whose induced transport path best matches the desired SOT objective. Thus, the GH-PID construction plays the role of a structured variational ansatz for SOT. In this section, we introduce and derive the GH-PID components that will serve as the foundation for this path-fitting framework.

\subsection{Controlled SDE and Path Integral}
\label{subsec:sde_path}

We consider a controlled Itô SODE (\ref{eq:sde}) with the terminal density law prescribed by Eq.~(\ref{eq:target}). In path-integral form, the law of a trajectory $x_{0\to 1}$ under a drift $u$ is
\begin{equation}
\label{eq:path_density}
p\!\left(x_{0\to 1}\mid u_{0\to 1}(\cdot)\right) \;\propto\; 
\exp\!\left(-\frac{1}{2}\int_0^1 \!\!\big\|\dot x_t - u_t(x_t)\big\|^2\,dt\right),
\end{equation}
and the time-$t$ marginal $p_{X_t}$ is obtained by integrating \eqref{eq:path_density} over all paths restricted to their value at $t$.

\subsection{Guided Green Functions}
\label{sec:guided_GF}

Associated to \eqref{eq:guided_potential} are the forward/backward Green functions $G_t^{(\pm)}$ solving the pair of linear -- backward and forward -- Kolmogorov--Fokker--Planck (KFP) equations
\begin{align}
\label{eq:kfp_backward}
-\partial_t G_t^{(-)}(x\mid y) \;+\; V_t(x)\,G_t^{(-)}(x\mid y) \;&=\; \frac{1}{2}\,\Delta_x G_t^{(-)}(x\mid y), 
& G^{(-)}_{t=1}(x\mid y) &= \delta(x-y),\\
\label{eq:kfp_forward}
\phantom{-}\partial_t G_t^{(+)}(y\mid 0) \;+\; V_t(y)\,G_t^{(+)}(y\mid 0) \;&=\; \frac{1}{2}\,\Delta_y G_t^{(+)}(y\mid 0),
& G^{(+)}_{t=0}(y\mid 0) &= \delta(y).
\end{align}
For the quadratic $V_t$ in \eqref{eq:guided_potential}, $G_t^{(\pm)}$ are Gaussian kernels whose quadratic and linear coefficients obey a coupled Riccati/linear ODE system detailed in Appendix~\ref{sec:Riccati}.

\subsection{Mixed Integrable Stochastic Optimal Transport}
\label{subsec:mixed_sot}

We formulate sampling as a \emph{Mixed} and \emph{Integrable} SOT problem with a \emph{hard} terminal constraint and \emph{soft} path guidance. In the low-integrability regime (no gauge field, no exogenous drift), the linearly solvable objective reads
\begin{equation}
\label{eq:sot_objective}
\min_{u}\;\; \mathbb{E}\!\left[\int_0^1 \!\Big(\tfrac{1}{2}\|u_t(x_t)\|^2 + \frac{\beta_t}{2}\,\big\|x_t-\nu_t\big\|^2\Big)\,dt\right]
\quad \text{s.t. Eqs.~(\ref{eq:sde},\ref{eq:target}).}
\end{equation}
For any fixed protocol $\Gamma_{0\to 1}=\{\beta_{0\to 1},\nu_{0\to 1}\}$ with $\nu_0=0$, the optimal control solving Eq.~(\ref{eq:sot_objective}) admits the closed form
\begin{equation}
\label{eq:score_from_probe}
u_t^\ast(x;\Gamma_{0\to 1}) \;=\; \nabla_x \log Z(t;x;\Gamma_{0\to 1}),
\qquad 
Z(t;x;\Gamma_{0\to 1}) \;\doteq\; \int_{\mathbb{R}^d} q (y\!\mid\! t;x;\Gamma_{0\to 1})\,dy,
\end{equation}
where
\begin{align}
\label{eq:probe_q}
q (y\mid t;x;\Gamma_{0\to 1}) &\;\doteq\; p^{(\tar)}(y)\,\exp\!\big(-\Delta(t;x;y;\Gamma_{0\to 1})\big),\\[2pt]
\label{eq:Delta_def}
\Delta(t;x;y;\Gamma_{0\to 1}) &\;\doteq\; -\log\frac{G_t^{(-)}(x\mid y;\Gamma_{0\to 1})}{G_1^{(+)}(y\!\mid\! 0;\Gamma_{0\to 1})} \;+\; C(t;x;\Gamma_{0\to 1}),
\end{align}
and the scalar $C(t;x;\Gamma_{0\to 1})$ is chosen so that $\int \exp(-\Delta)\,dy = 1$. The corresponding (normalized) \emph{probe density} is
\begin{equation}
\label{eq:probe_density}
p (y\!\mid\! t;x;\Gamma_{0\to 1}) \;=\; \frac{q (y\!\mid\! t;x;\Gamma_{0\to 1})}{Z (t;x;\Gamma_{0\to 1})}.
\end{equation}
Two useful limiting relations are
\begin{equation}
\label{eq:probe_limits}
\lim_{t\to 1^-} p (y\!\mid\! t;x;\Gamma_{0\to 1}) = \delta(x-y),
\qquad
\lim_{t\to 0^+} p (y\!\mid\! t;x;\Gamma_{0\to 1}) = p^{(\tar)}(y).
\end{equation}

For later use, Appendix~\ref{sec:yhat} introduces the \emph{predicted final state} map
$\hat y(t;x;\Gamma_{0\to 1})$ and shows that \eqref{eq:score_from_probe} results in Eq.~(\ref{eq:ustar_compact_forwardref}) --- which we restate here for convenience
\begin{equation*}
u_t^\ast(x;\Gamma_{0\to 1}) \;=\; b_t^{(-)}\!\left(\hat y(t;x;\Gamma_{0\to 1})\;-\; \mu_t(x)\right),
\end{equation*}
where $b_t^{(-)}$ and the reweighting mean $\mu_t(x)$ (affine in $x$) are determined by the Riccati coefficients of Appendix~\ref{sec:Riccati} (see Eqs.~\eqref{eq:K_def}--\eqref{eq:mu_rho}) \footnote{Throughout the paper -- including the appendices -- we simplify notation by omitting the explicit dependence on the protocol $\Gamma$ in quantities such as $b_t^{(-)}$ and $\mu_t(x)$ (i.e., we write $b_t^{(-)}$, $\mu_t(x)$ instead of $b_t^{(-)}(\Gamma_{0\to 1})$, $\mu_t(x;\Gamma_{0\to 1})$). This dependence should be understood from context and will be reinstated when necessary.}.
The representation \eqref{eq:ustar_compact_forwardref} will be useful for analysis and numerical experiments. 

\paragraph{Why this formulation, and why now?} The machinery above (guided potential \eqref{eq:guided_potential}, linearly solvable objective \eqref{eq:sot_objective}, and closed forms for $u_t^\ast$ via \eqref{eq:score_from_probe}, eventually resolved due to integrability into \eqref{eq:ustar_compact_forwardref}) formally pins down the problem. Yet one might ask: \emph{if the terminal law is fixed, why care about the path?} The answer is operational: partial trajectories often carry value (anytime use), intermediate checkpoints feed downstream tasks, numerical budgets favor well-conditioned flows, and safety/preferences are naturally expressed along the way. Crucially, GH-PID lets us encode these path preferences \emph{without} giving up solvability -- simply by shaping the protocol $\Gamma_t=\{\beta_t,\nu_t\}$. In the next section, we show how to cast such needs as \emph{mixed} -- and more general than H-PID -- SOT problems with hard terminal matching and soft guidance, and we give practical design patterns for $\Gamma_t$ that remain analytic, explainable and computationally stable.

\section{Diagnostics for Protocol Comparison and Learning}
\label{sec:diagnostics}

Guided Harmonic PID (GH--PID) produces stochastic trajectories $\{x_t\}_{t\in[0,1]}$ by driving the diffusion \eqref{eq:sde} with the closed-form optimal drift $u_t^\ast$ solving Eq.~(\ref{eq:principle-SOT}). Given a protocol 
\[
\Gamma_{0\to1}=\bigl\{(\beta_t,\nu_t)\bigr\}_{t\in[0,1]},
\]
our goal in the following Sections~\ref{sec:navigation} and \ref{sec:cases}  is to compare different protocols and to \emph{learn} improved ones. For this purpose we use a small set of diagnostics that quantify precisely the features needed in the navigation experiments of Section~\ref{sec:cases}:   adherence to the guide, geometric shaping of the ensemble, and terminal accuracy.  

\vspace{1ex}
\subsection{Integrated guide cost}

Throughout the navigation experiments the guidance cost is the quadratic potential (\ref{eq:guided_potential}), so that the protocol penalizes deviations from the centerline. The associated integrated cost -- that is optimization objective in Eq.~(\ref{eq:principle-SOT}) -- is
\begin{equation}
J^{\rm (g)}(\Gamma)
\;=\;\int_0^1 \mathbb{E}\!\left[V_t(x_t)\right] dt\;=\;\int_0^1 \frac{\bar{\beta}_t}{2}\,\mathbb{E}\!\left[\|x_t-\bar{\nu}_t\|^2\right]\,dt,
\label{eq:Jguide}
\end{equation}
with $x_t$ evolving under GH--PID. 

Notice, the difference between the guide cost (\ref{eq:Jguide}) and the PID cost, which one gets substituting Eq.~(\ref{eq:guided_potential}) in Eq.~(\ref{eq:principle-SOT}).

$J^{\rm (g)}(\Gamma)$ is used in Sections~\ref{subsec:caseA},\ref{subsec:cases_caseA} to compare hand-crafted protocols, in Sections~\ref{subsec:caseB},\ref{subsec:caseBresults} as the main optimization objective, and in Sections~\ref{subsec:caseC},\ref{subsec:caseCresults} as the building block for multi-expert fusion.

\vspace{1ex}
\subsection{Time-resolved adherence}

In addition to the aggregate cost \eqref{eq:Jguide}, it is often informative to
examine its instantaneous contribution,
\begin{equation}
A(t)=V_t(x_t)= \frac{\beta_t}{2}\,
  \mathbb{E}\!\left[\|x_t-\nu_t\|^2\right].
\label{eq:AofT}
\end{equation}
Adherence curves $A(t)$ are used repeatedly in Sections~\ref{subsec:caseA}--\ref{subsec:caseB} and \ref{subsec:cases_caseA}--\ref{subsec:caseBresults} to interpret curvature-induced lag, confinement effects from large~$\beta_t$, and path-shaping differences between protocols.

\subsection{Terminal fidelity}

A core property of GH--PID is that, for any protocol $\Gamma_t$, the terminal
distribution is exactly the target:
\[
x_1 \sim p_{\rm tar}.
\]
In all navigation experiments we verify this property empirically by comparing the final-time ensemble with the prescribed GMM target or its product-of-experts fusion (Sections~\ref{subsec:caseC},\ref{subsec:caseCresults}).  This check confirms that protocol comparisons and protocol learning alter only the \emph{path geometry}, never the hard terminal constraint.

\paragraph{Summary.} The navigation studies in Sections~\ref{sec:navigation}--\ref{sec:cases} rely on the three diagnostics above: (i)~integrated guide cost $J_{\rm guide}$, (ii)~time-resolved adherence $A(t)$, and (iii)~terminal fidelity. Together they provide the minimal information necessary to evaluate and learn guidance protocols in the GH--PID framework.

\section{Navigation as Mixed Stochastic Optimal Transport}
\label{sec:navigation}

In this section we specialize Guided Harmonic PID (GH--PID) to a stylized autonomous navigation problem in the plane.  The environment is represented by a ``soft corridor'' that connects an entry region at the origin to a terminal region where the hard constraint $p^{(\mathrm{tar})}$ is supported.  A protocol $\Gamma_{0\to 1}=\{(\nu_t,\beta_t)\}_{t\in[0,1]}$ specifies both a moving centerline $\nu_t\in\mathbb{R}^2$ and a time--dependent stiffness $\beta_t>0$ of a quadratic guide potential.  The GH--PID sampler then generates controlled paths $\{x_t\}_{t\in[0,1]}$ that (i) end in $p^{(\mathrm{tar})}$ and (ii) are softly encouraged by the diagnostics of Section~\ref{sec:diagnostics} to stay inside a low--risk tube around $\nu_t$ while traversing the corridor.

We use this setting to view GH--PID as an empirical solver of the mixed SOT problem~\eqref{eq:principle-SOT}: the hard terminal constraint pins the final distribution, whereas the protocol $\Gamma$ is designed so as to sculpt and, in later sections, optimize integrated path costs.  Our focus here is on how \emph{geometric features of the corridor} are encoded directly into $\Gamma_t=(\nu_t,\beta_t)$, rather than into an external cost functional or a learned neural controller.

We work in $d=2$ with a fixed pair of Gaussian -- mixture (GMM) terminal densities $p^{(\mathrm{tar})}$: a two--mode GMM elongated along the corridor, and a three--mode GMM in a triangular configuration.  All particles are initialized at the origin $x_0=0$, and the guide $\nu_t$ is constructed so that $\nu_0=0$ and $\nu_1$ coincides with the center of mass of the two--mode GMM. Thus all protocols share the same start and end points but differ in how they traverse the interior of the corridor.

We present three navigation cases of increasing sophistication.  They share the same underlying corridor endpoints and terminal laws but differ in how the protocol is constructed and, in the latter two cases, how it is used for empirical SOT fitting.  Case~A uses fully hand--crafted continuous centerlines and stiffness profiles to define a small family of protocols that illustrate three key knobs: \emph{corridor geometry}, \emph{confinement level}, and \emph{temporal progression} along the guide.  Case~B introduces a low -- dimensional parametric family of such protocols and learns the parameters from data by minimizing a diagnostic objective.  Case~C considers multi--task navigation, where several terminal objectives and landscapes are fused into a single product--of--experts terminal distribution.  Appendix~\ref{app:navigation} provides the explicit centerline and stiffness templates, the piecewise--constant (PWC) discretized protocols used in the sampler, and additional implementation details.

\subsection{Case A: Hand-Crafted Protocols in a 2D Corridor}
\label{subsec:caseA}

We begin with a purely constructive scenario that isolates the three main degrees of freedom in GH--PID navigation \footnote{Since Case A involves diagnosis of a fixed protocol $\Gamma_t$ -- rather than optimization of an expert or desiderata protocol $\bar{\Gamma}_t$, as in Cases B and C -- we do not maintain a distinction between $\Gamma_t$ and $\bar{\Gamma}_t$ here, and use the unbarred notation throughout.}:
\begin{enumerate}
  \item the \emph{geometry} of the guide $\nu_t$ (straight vs.\ V--shaped vs.\ S--shaped),
  \item the \emph{strength} of confinement via $\beta_t$, and
  \item the \emph{temporal progression} along a given geometric guide.
\end{enumerate}

All experiments in Case~A share the following common setting.

\paragraph{Intrinsic corridor frame.}
We connect the origin $x_{\mathrm{in}}=0\in\mathbb{R}^2$ to the center of mass $x_{\mathrm{out}}\in\mathbb{R}^2$ of the two--mode GMM by a straight segment.  Let
\[
  v = x_{\mathrm{out}} - x_{\mathrm{in}},\qquad
  e = \frac{v}{\|v\|},\qquad
  n = (-e_2,e_1),
\]
denote respectively the axis direction, its unit vector, and a unit normal. A scalar parameter $s\in[0,1]$ measures progress along the axis, $x_{\mathrm{axis}}(s)=x_{\mathrm{in}}+s\,v$, while transverse displacements are taken along~$n$.  All centerlines are of the form
\begin{equation}
  \nu(s) = x_{\mathrm{in}} + s\,v + \Delta(s)\,n,\qquad s\in[0,1],
  \label{eq:centerline_generic}
\end{equation}
with different choices of a scalar offset $\Delta(s)$.

\paragraph{Straight, V-- and S--centerlines.} We consider three continuous templates:
\begin{align}
  \text{Straight:}\quad
  &\Delta_{\mathrm{lin}}(s) \doteq 0,\\[0.25em]
  \text{V--neck:}\quad
  &\Delta_{\mathrm{V}}(s) = -A_{\mathrm{V}}\,
      \bigl(1 - |2s-1|\bigr),\\[0.25em]
  \text{S--tunnel:}\quad
  &\Delta_{\mathrm{S}}(s) = A_{\mathrm{S}}\,
      \sin(2\pi s),
\end{align}
with amplitudes $A_{\mathrm{V}},A_{\mathrm{S}}>0$ chosen so that the maximal transverse excursion satisfies $|\Delta(s)|\approx 1.5$ in the V-- and S--cases. The straight centerline corresponds to a direct traversal from entry to target, while the V--neck and S--tunnel induce, respectively, a single deep excursion and a sinusoidal S--shaped path within the same corridor.

In all experiments we identify the diffusion time $t\in[0,1]$ with the centerline parameter $s\in[0,1]$ and write $\nu_t=\nu(s{=}t)$, except in Case~A3 where the progression is deliberately truncated.

\paragraph{PWC protocols.} The continuous pair $(\nu_t,\beta_t)$ is approximated by a PWC protocol $\Gamma^{(\mathrm{  PWC})}=\{(\nu_k,\beta_k)\}_{k=0}^{K-1}$ on a partition $0=t_0<\dots<t_K=1$.  To ensure that both endpoints are exactly represented by the discrete protocol we set
\begin{equation}
  \nu_0 = \nu_0^{\phantom{\star}},\qquad
  \nu_{K-1} = \nu_1^{\phantom{\star}},
\end{equation}
and define the remaining $\nu_k$ at midpoints $t_k^\star=\tfrac12(t_k+t_{k+1})$ for $1\le k\le K-2$.  The stiffness is sampled at midpoints for all segments,
\begin{equation}
  \beta_k = \beta_{t_k^\star},\qquad k=0,\dots,K-1.
\end{equation}
The resulting PWC schedule is passed to the analytic GH--PID machinery (via the GuidedPWCSchedule object) of Section~\ref{sec:formulation}.  The discrete $\nu_k$ appear as markers in the figures, while the continuous centerlines are shown as dashed curves.

Within this framework we construct three illustrative families of protocols.

\paragraph{Case A1: Geometry at fixed stiffness.} The first experiment isolates the effect of geometry by fixing a constant stiffness profile, $\beta_t >0$, and varying only the centerline template in~\eqref{eq:centerline_generic}. We build three protocols $\Gamma^{(\mathrm{lin})}, \Gamma^{(\mathrm{V})}, \Gamma^{(\mathrm{S})}$, corresponding to the straight, V--neck and S--tunnel centerlines, all with the same $\beta_{\mathrm{geom}}$ and the same start and end points.  The GH--PID sampler produces path ensembles that are compared across geometries using the diagnostics of Section~\ref{sec:diagnostics}: adherence to the guide (mean distance to $\nu_t$), dispersion (empirical covariance), and control effort $\|u_t^\ast(x_t;\Gamma)\|^2$.

\paragraph{Case A2: Confinement at fixed S--geometry.}

The second experiment fixes the S--tunnel geometry $\nu_t=\nu^{(\mathrm{S})}_t$ and varies only the confinement strength via a constant--in--time but scaled stiffness,
\[
  \beta_t^{(\gamma)} \doteq  \gamma\,\beta_{\mathrm{base}},
  \qquad \gamma\in\{\gamma_1,\dots,\gamma_L\},
\]
with $\beta_{\mathrm{base}}>0$ fixed and $\gamma$ ranging from loose to tight confinement.  This produces a family of protocols $\Gamma^{(\mathrm{S},\gamma)}_t=(\nu^{(\mathrm{S})}_t,\beta_t^{(\gamma)})$. The resulting path ensembles illustrate how increasing $\gamma$ squeezes the cloud toward the centerline, reducing lateral dispersion and control variability while preserving terminal sampling.

\paragraph{Case A3: Temporal progression along the V--neck.}

The third experiment fixes both the V--neck geometry and the stiffness level but varies the temporal progression along the V--shaped guide.  We define
\[
  \nu^{(\mathrm{V},s_{\max})}_t
    = \nu^{(\mathrm{V})}_{\min(t,s_{\max})},\qquad
    s_{\max}\in\{1.0,0.5,0.0\},
\]
so that the guide moves along the same V--centerline as in Case~A1 but stops early at a fraction $s_{\max}$ of the total arclength and remains frozen there for the remainder of the diffusion.  The stiffness is held constant, $\beta_t\doteq\beta_{\mathrm{V}}$.

For $s_{\max}=1$ the guide traverses the entire V--neck; progressively smaller $s_{\max}$ produce protocols that linger in earlier parts of the corridor and hand over more of the transport burden to the GH--PID drift and noise.  The corresponding diagnostics reveal how temporal scheduling of the guide affects the speed and reliability with which the cloud reaches the target region.

Case~A involves no optimization: all amplitudes, stiffness levels, and cutoff fractions are chosen by hand.  Its purpose is to illustrate, in a transparent and fully controllable setting, how geometric and temporal features of $\Gamma_t=(\nu_t,\beta_t)$ are reflected in empirically measured diagnostics and in the spatial evolution of the cloud.  The resulting figures are summarized in Section~\ref{sec:cases} and detailed formulas appear in Appendix~\ref{app:caseA}.

\subsection{Case B: Protocol Learning for a Single Navigation Task}
\label{subsec:caseB}

\paragraph{Stage reset.} In Case~A we explored fully hand--crafted protocols $\Gamma_t = (\nu_t,\beta_t)$ and observed three recurring phenomena: (i) the \emph{geometry} of the centerline $\nu_t$ controls lag, curvature--induced deviations, and the timing of modal splitting; (ii) the \emph{stiffness} $\beta_t$ governs confinement, exploration, and numerical conditioning; and (iii) the \emph{temporal progression} of the guide acts as an independent lever that biases mass toward modes closest to the final guide position.  For Case~B we deliberately focus on a simpler, more controlled setting: a single two--mode navigation task in the same corridor, with a fixed stiffness profile and a prescribed ``desiderata'' centerline, and we optimize only the guidance $\nu_t$ on a piecewise--constant (PWC) grid via automatic differentiation.

\paragraph{Desiderata protocol and PWC parametrization.}
We start from a continuous expert/desiderata protocol
\[
\bar\Gamma_t=(\bar\nu_t,\bar\beta_t),
\qquad t\in[0,1],
\]
constructed as follows.  The expert/desiderata centerline $\bar\nu_t$ is an S--shaped curve in the intrinsic corridor frame:
\[
\nu^{\rm (raw)}(t) = x^{\rm (axis)}(t) + A^{\rm (swing)}\,\tanh\!\bigl(\kappa(2t-1)\bigr)\,n^{\rm (axis)},
\]
where $x^{\rm (axis)}(t)$ is the straight segment from $x_0$ to $x_1$ and $n^{\rm (axis)}$ is the transverse unit normal.  A linear endpoint correction enforces
\[
\bar\nu_t
=
\nu^{\rm (raw)}(t)
+ (1-t)\bigl(x_0-\nu^{\rm (raw)}(0)\bigr)
+ t\bigl(x_1-\nu^{\rm (raw)}(1)\bigr),
\]
so that $\bar\nu_0=x_0=0$ and $\bar\nu_1=x_1$.  The expert/desiderata stiffness is constant,
\[
\bar\beta_t \doteq \beta^{\rm (const)}>0,
\]
and is chosen from the stable regime identified in Case~A.

For GH--PID sampling we discretize $[0,1]$ into $K^{\rm (pwc)}$ intervals with endpoints
\[
0=t_0<t_1<\dots<t_{K^{\rm (pwc)}}=1,
\]
and represent the learnable protocol as a PWC guide
\[
\Gamma(t;\theta) = (\nu(t;\theta),\bar\beta_t),\qquad
\nu(t;\theta) \doteq \nu_k(\theta)\ \text{for}\ t\in[t_k,t_{k+1}),
\]
where $\theta=\{\nu_k\}_{k=0}^{K^{\rm (pwc)}-1}$ collects the midpoint values.  We fix $\nu_0=x_0$, leave $\nu_{K^{\rm (pwc)}-1}$ unconstrained, and initialize all $\nu_k$ at the expert/desiderata midpoints $\bar\nu(t_k^{\rm (mid)})$, $t_k^{\rm (mid)}=\tfrac{1}{2}(t_k+t_{k+1})$.  This PWC parametrization preserves the analytic GH--PID machinery: the Green--function coefficients and the optimal drift $u_t^*(x;\Gamma)$ are obtained from the same Riccati/linear updates as in Case~A, but now driven by a trainable centerline.

\paragraph{Expert/Desiderata--guided objective.} Given a protocol $\Gamma(t;\theta)$, we simulate GH--PID paths via the controlled SDE~\eqref{eq:sde} with drift $u_t^*(x;\Gamma)$, starting from $x_0=0$ and targeting the fixed two-- (or three-) mode GMM of Section~\ref{sec:navigation}.  The expert/desiderata protocol $\bar\Gamma_t$ defines a path--$\nu_t$--dependent cost that penalizes deviation from the desired (e.g. S--shaped) tube (of width $\beta_t$):
\begin{equation}
  J^{\rm (des)}(\theta)
  = \int_0^1 \frac{\bar\beta_t}{2}\,
    \mathbb{E}\bigl[\left(x_t-\bar\nu_t\right)^2\bigr]\,dt.
  \label{eq:caseB_desiderata_cost}
\end{equation}
Let us emphasize that $\Gamma_t$ is not equal to $\bar{Gamma}_t$ -- the latter is fixed,  while the former is yet to be determined by optimizing a regularized version of Eq.~(\ref{eq:caseB_desiderata_cost}). Also the "desiderata" cost (\ref{eq:caseB_desiderata_cost}) is not equal to the GH-PID optimal cost-to-go -- given by Eq.~(\ref{eq:principle-SOT}) with $u\to u^*$ and $C\to C^{\rm (PID)}$ -- the difference is in the additional $u^2$-term in the integrand of the latter. 

To maintain terminal fidelity we add a regularizing soft cross--entropy term at $t=1$,
\begin{equation}
  J^{\rm (CE)}(\theta)
  = -\mathbb{E}\bigl[\log p^{\rm (tar)}(x_1)\bigr],
  \label{eq:caseB_CE_cost}
\end{equation}
where $p^{\rm (tar)}$ is the target GMM density, and form
\begin{equation}
  J^{\rm (state)}(\theta)
  = J^{\rm (des)}(\theta)
  + \lambda^{\rm (CE)}\,J^{\rm (CE)}(\theta),
\end{equation}
and then include a mild quadratic regularizer that discourages wild excursions away from the desiderata:
\begin{equation}
  J^{\rm (reg)}(\theta)
  = \lambda_\nu\sum_{k=0}^{K^{\rm (pwc)}-1}
    \|\nu_k-\bar\nu(t_k^{\rm (mid)})\|^2.
\end{equation}
The total Case~B optimization objective is
\begin{equation}
  J(\theta) = J^{\rm (state)}(\theta)+J^{\rm (reg)}(\theta),
  \label{eq:caseBobjective}
\end{equation}
which trades off adherence to the guidance, terminal likelihood, and smooth deviations from the expert/desiderata centerline.

\paragraph{Autograd implementation.} The GH--PID sampler with PWC protocol is differentiable with respect to the centerline parameters $\theta=\{\nu_k\}$.  We implement the full pipeline in PyTorch: starting from $\theta=\bar\theta$ (desiderata midpoints), we simulate $M=4000$ paths with an Euler--Maruyama discretization of length $T$ and accumulate the Monte--Carlo estimate of $J(\theta)$ in~\eqref{eq:caseBobjective}.  Automatic differentiation provides stochastic gradients $\nabla_\theta J(\theta)$, and we update $\theta$ using Adam with an adaptive learning rate (reduced whenever $J$ fails to improve for several iterations).  After convergence we obtain an optimized PWC protocol $\Gamma^*_t=(\nu_t^*,\bar\beta_t)$, which is then analyzed and compared to the expert/desiderata and to a straight--axis baseline in Section~\ref{subsec:caseBresults}.

\subsection{Case C: Multi-Task Navigation via Consensus Fusion of Experts}
\label{subsec:caseC}

\paragraph{Motivation and stage setting.} Cases~A and~B demonstrated that (i) \emph{trajectory geometry matters} -- the shape of $\nu_t$ strongly affects lag, curvature-induced drift, and mode-splitting; and (ii) modest learning of a PWC guidance protocol can substantially improve adherence while preserving exact terminal sampling.  Case~C builds on these insights by introducing a more realistic scenario involving multiple, internally coherent experts that disagree on both \emph{terminal beliefs} and \emph{trajectory-level preferences}.

\begin{quote}
\emph{Given two experts whose recommendations may be somehow in a conflict at both the destination and the path level, can a \textbf{single learned protocol} reconcile their views in a principled, interpretable, and analytically tractable way?}
\end{quote}

To produce a sharp and visually compelling test, we deliberately restrict to a focused, high-contrast configuration that isolates GH--PID’s ability to form a negotiated consensus.

\paragraph{Experts with terminal beliefs and trajectory preferences.} Each expert $m\in\{1,2\}$ supplies a \emph{complete hypothesis} consisting of:
\begin{enumerate}
\item A terminal belief $p_m^{\rm (tar)}(x)$, given as a GMM.  The expert’s preferred terminal guidance point is the  mean of its target distribution   $x_{m}^{\rm (out)} = \mathbb{E}_{p_m^{\rm (tar)}}[x]$.

\item An expert curve $\bar\nu^{(m)}_t$, constructed to connect the common entry point $x(0)=0$ with $x_{m}^{\rm (out)}$ and to encode the expert’s preferred geometric style of navigation (e.g., direct, curved, cautious, exploratory).
\end{enumerate}

Thus the two experts may disagree about both \textbf{where} the swarm should end
and \textbf{how} it should move through the corridor.

\paragraph{Commander fusion: exact product-of-experts terminal laws.} Here we follow the classic product-of-expert methodology \cite{hinton_training_2002}. The commander expresses relative trust in the two experts through weights $(\alpha_1,\alpha_2)\in\{(1,1),(2,1),(1,2)\}$.   We restrict attention to the following \emph{three exact GMMs}:
\begin{align}\label{eq:three-prod-GMM}
p^{\rm (tar)}_{(1,1)}(x) & \propto p^{\rm (tar)}_1(x)\,p^{\rm (tar)}_2(x),\quad
p^{\rm (tar)}_{(2,1)}(x) \propto (p^{\rm (tar)}_1(x))^2\,p^{\rm (tar)}_2(x),\\
p^{\rm (tar)}_{(1,2)}(x) & \propto p^{\rm (tar)}_1(x)\,(p^{\rm (tar)}_2(x))^2.
\end{align}
Products of Gaussian mixtures (experts) remain Gaussian mixtures (with appropriately reweighted components and precisions), so all three fused targets are \emph{exact} GMMs and thus fully compatible with GH--PID’s analytic formulas for $\hat{y}(t;x)$ and $u_t^*(x)$. These three models correspond to: equal trust, trust favoring expert~1, and trust favoring expert~2.

\paragraph{Commander fusion: trajectory-level compromise.} The same trust choices determine how strongly each expert influences the path geometry. Each expert contributes a soft trajectory cost
\[V^{(m)}_t(x)=\frac{\beta_t}{2}\left(x-\nu^{(m)}_t\right)^2,\]
and the commander seeks a \emph{single} PWC protocol $\Gamma_t=(\nu_t,\beta_t)$ minimizing
\begin{equation}
\mathcal{J}^{\rm (multi)}(\Gamma)
  = \alpha_1 \!\!\int_0^1 \mathbb{E}[V^{(1)}_t(x_t)]\,dt
  + \alpha_2 \!\!\int_0^1 \mathbb{E}[V^{(2)}_t(x_t)]\,dt.
\label{eq:J-multi}
\end{equation}
By using the same $(\alpha_1,\alpha_2)$ at the terminal and path levels, the commander fuses each expert’s hypothesis consistently and transparently.

\paragraph{Learning and expected outcomes.} Case~C now proceeds as follows:
\begin{enumerate}
\item Construct two geometrically distinct expert curves $\bar\nu^{(1)}_t$ and $\bar\nu^{(2)}_t$ with different endpoints.
\item Form one of the three exact fused GMMs $p^{\rm (tar)}_{(1,1)}$, $p^{\rm (tar)}_{(2,1)}$, $p^{\rm (tar)}_{(1,2)}$ depending on the commander’s trust.

\item Use autograd to learn a single compromise centerline $\nu_t$ minimizing $\mathcal{J}^{\rm (multi)}(\Gamma)$.
\item Evaluate how the optimized GH--PID sampler:
      \begin{itemize}
      \item balances the two trajectory geometries,
      \item steers samples toward the fused terminal distribution, and
      \item modifies its behavior under changes in commander trust.
      \end{itemize}
\end{enumerate}

This setting -- with some additional regularization -- yields a clear “\emph{consensus-from-conflict}” demonstration:  GH--PID computes, in closed form, the negotiated compromise between diverging expert recommendations --- at both the trajectory and terminal levels --- and learns a single protocol that faithfully carries out the commander’s fused intent. Results appear in Section~\ref{subsec:caseCresults}.

\section{Case Studies}\label{sec:cases}

\subsection{Case A: 2D Corridor Experiments}
\label{subsec:cases_caseA}

We first illustrate GH--PID as an empirical solver of mixed SOT in the
hand--crafted 2D corridor setting of Section~\ref{subsec:caseA}.  All path
ensembles in this subsection start at the origin, end in the prescribed GMM
targets, and are driven by analytic GH--PID drifts under PWC protocols
$\Gamma^{(\mathrm{  PWC})}$ constructed from continuous templates
$(\nu_t,\beta_t)$ as in Appendix~\ref{app:caseA}.  We visualize both the guide
(centerline and current $\nu_t$) and the empirical cloud via snapshots at ten
intermediate times $t\in[0,1]$, including $t=0^+$.

\paragraph{Geometry flexibility (Case A1).}

\begin{figure}[t]
  \centering
  \includegraphics[width=\textwidth]{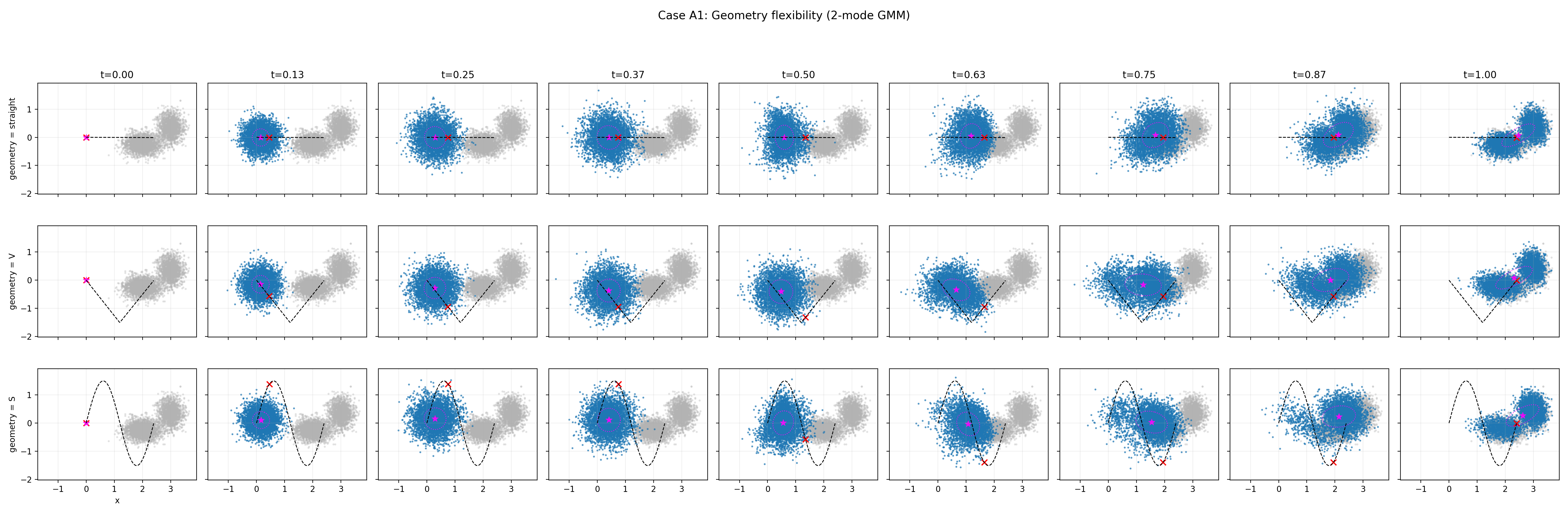}
  \includegraphics[width=\textwidth]{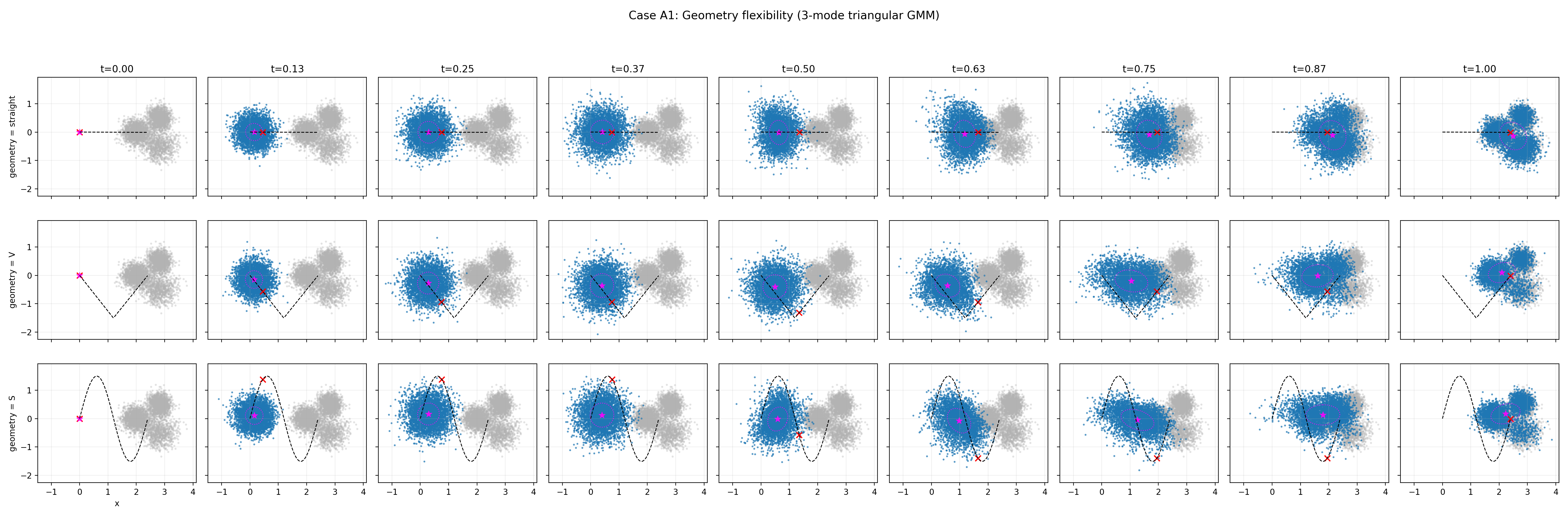}
  \caption{Case~A1: geometry flexibility at fixed stiffness for 2 mode (top) and 3 mode (bottom) examples. Each row corresponds to a different guide geometry (straight, V--neck, S--tunnel), and columns show snapshots at increasing times $t\in[0,1]$.  Gray dots: samples from the terminal two--mode GMM. Colored dots: GH--PID cloud at time $t$.  Black dashed curve: analytic centerline $\nu_t$.  Red cross: current guide location $\nu_t$.  Magenta star and dotted ellipse: empirical mean and covariance of the cloud. GH--PID faithfully reaches the terminal law while expressing markedly different path geometries.}
  \label{fig:caseA_geometry}
\end{figure}

Fig.~\ref{fig:caseA_geometry} compares three protocols that share the same terminal distributions and a constant stiffness $\beta_t\doteq \beta_{\mathrm{geom}}$ but differ in their centerlines: straight, V--neck, and S--tunnel.  Each row of panels corresponds to one geometry, and columns show successive time slices. In every panel we plot: (i) samples from the terminal GMM (gray), (ii) the current cloud $\{x_t\}$ (colored dots), (iii) the analytic centerline (black dashed curve), (iv) the current guide location $\nu_t$ (red cross), and (v) the empirical mean and covariance of the cloud (magenta star and dotted ellipse).

The straight guide produces nearly ballistic transport along the axis, with the cloud remaining broadly dispersed around the centerline.  The V--neck guide pulls the cloud into a deep transverse excursion and then returns it to the terminal region, while the S--tunnel induces two bends that the cloud follows closely.  Despite these geometric differences, the terminal sampling remains faithful to the GMM targets, illustrating how GH--PID can accommodate qualitatively different paths at fixed hard constraint.

The Monte--Carlo trajectories generated by GH--PID reflect both the geometry of the guide $(\nu_t)$ and the stiffness of the confining potential $\beta_t$.  Because the sampler transports the cloud from a highly localized initial condition to a multi--modal terminal GMM, the evolution of empiricalmeans, covariances, and mode weights makes visible several qualitative phenomena:
\begin{itemize}
\item \textbf{Lag behind the guide.} Even when the drift uses the analytic control $u_t^\ast$, the empirical cloud generally trails behind the moving centerline $\nu_t$.  This time--lag is a combined effect of stochasticity, finite stiffness $\beta_t$, and discretization.  It is most noticeable when $\nu_t$ bends abruptly or accelerates.

\item \textbf{Early modal splitting.} When the geometric path of $\nu_t$ bends strongly (V-- or S--shape), the cloud tends to split into multiple lobes \emph{earlier} in time than in the straight--line case.   This is because the drift field induced by $u_t^\ast$ shifts probability mass toward different regions of $p_{\mathrm{tar}}$ depending on the local orientation of $\nu_t$, thus “previewing’’ the terminal modes.

\item \textbf{Effect of stiffness.}  Increasing $\beta_t$ tightens confinement around $\nu_t$, producing well--aligned tubes of trajectories but also amplifying numerical stiffness: the sampler becomes more sensitive to discretization, and small time--integration errors skew terminal mode weights.
\end{itemize}

\paragraph{Influence of geometric complexity of $\nu_t$.} Fig.~\ref{fig:caseA_geometry} compares three protocols that share the same stiffness profile $\beta(t)\doteq \beta_0$ but differ in their centerline geometry: a straight corridor, a V--shaped corridor, and an S--tunnel.

Two robust effects are visible:
\begin{enumerate}
\item \textbf{Lag between the guide and the cloud.} In all three geometries the red cross marks the instantaneous guide $\nu_t$ while the magenta star marks the empirical mean of the cloud. The mean consistently trails behind the guide, especially at times where $\nu_t$ bends.   The effect is weakest for the straight path, stronger for the V--neck, and strongest for the S--tunnel where multiple changes of curvature occur.

\item \textbf{Early appearance of modal structure.} For the V and S cases, the cloud begins to bifurcate into multiple lobes \emph{earlier} than along the straight path.   The turning of $\nu_t$ effectively “steers’’ different regions of the cloud toward different components of the terminal GMM.   This illustrates that GH--PID incorporates geometric information directly into the transport: convoluted motion of the guide induces anticipatory branching of the flow, visible well before $t=1$.
\end{enumerate}

These observations confirm that geometric features of $\nu_t$ have first--order effects on mode formation, trajectory allocation, and the instantaneous organization of probability mass.

\paragraph{Confinement flexibility (Case A2).}

\begin{figure}[t]
  \centering
  \includegraphics[width=\textwidth]{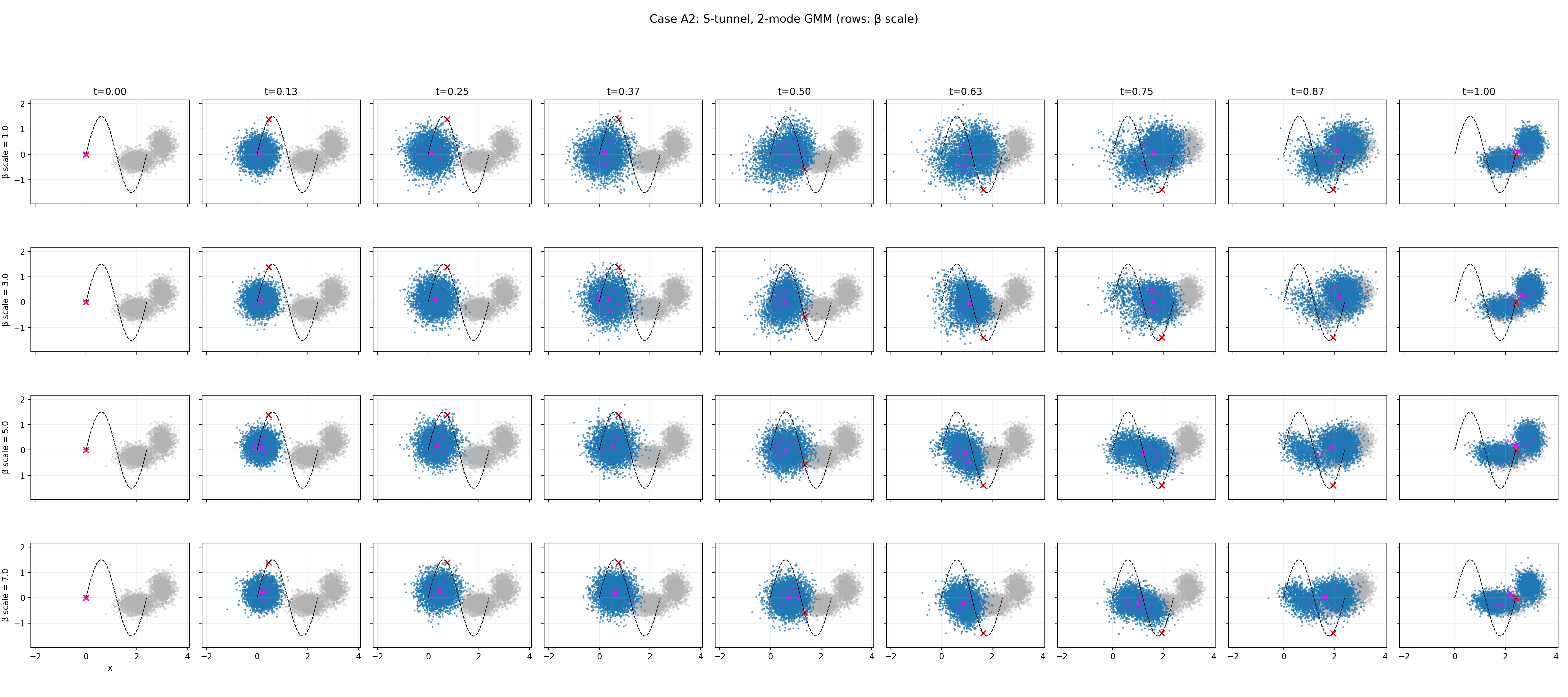}
  \includegraphics[width=\textwidth]{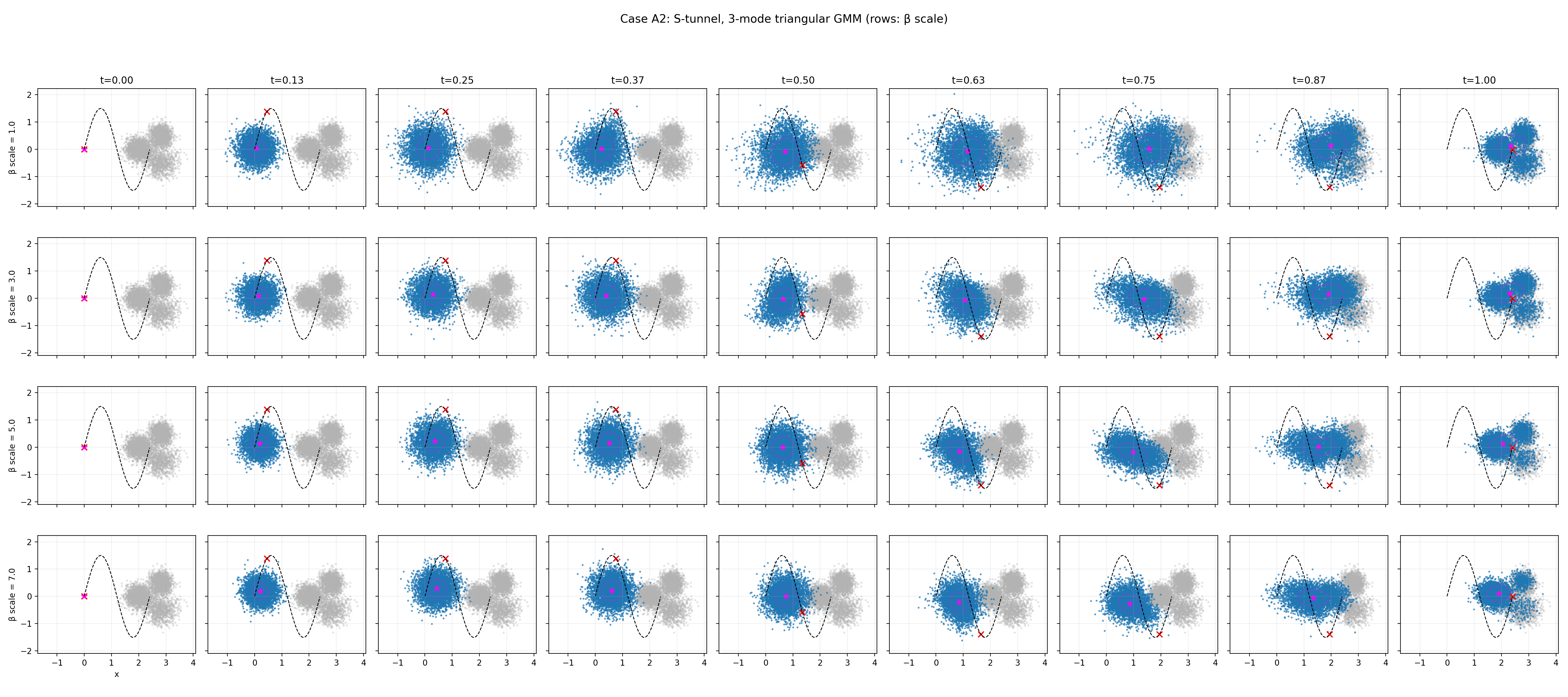}
  \caption{Case~A2: confinement flexibility along a fixed S--tunnel guide for 2 mode (top) and 3 mode (bottom) examples. Each row corresponds to a different stiffness scale $\beta_t^{(\gamma)}\doteq \gamma\,\beta_{\mathrm{base}}$, with $\gamma$ increasing from top to bottom.  Higher stiffness produces tighter confinement of the cloud around the centerline and reduced lateral variance, while preserving the terminal two--mode GMM.}
  \label{fig:caseA_A2}
\end{figure}

Fig.~\ref{fig:caseA_A2} explores the effect of stiffness on path confinement.  Here we fix the S--tunnel centerline $\nu^{(\mathrm{S})}_t$ and consider a family of constant--in--time stiffness profiles $\beta_t^{(\gamma)}\doteq \gamma\,\beta_{\mathrm{base}}$ with scale factors $\gamma$ ranging from loose to tight -- $\gamma \in \{1,5,10,30\}$.  Each row corresponds to a different $\gamma$, with the same visualization as above.

For small $\gamma$ the cloud diffuses broadly around the S--shaped guide and only gradually concentrates near the terminal GMM.  As $\gamma$ increases, the cloud becomes tightly confined to a narrow tube around the centerline, with markedly reduced lateral variance and a more coherent progression along the corridor.  This experiment highlights how the single scalar knob $\beta_t$ modulates the tradeoff between exploration and adherence, while the analytic GH--PID construction guarantees correct terminal sampling for all choices.

The following systematic trends appear:
\begin{enumerate}
\item \textbf{Confinement increases monotonically with $\beta$.} As $\gamma$ grows, the cloud becomes increasingly concentrated around the instantaneous guide.  The empirical covariance ellipsoids shrink, and trajectories deviate far less from $\nu_t$.

\item \textbf{Stiff dynamics amplify numerical effects.} Large $\beta$ makes the drift term $-a_t^{(-)}(x - \nu_t)$ extremely stiff. This causes two artifacts:
  \begin{itemize}
  \item The mean may overshoot or undershoot the guide at late times;
  \item Terminal mixture weights may deviate slightly from the analytic GMM, with over--representation of the component located closest to the end of the guide.
  \end{itemize}
  Both effects diminish with finer time discretization and larger particle ensembles, but are inherent to stiff guided diffusions under Euler--Maruyama.
\end{enumerate}

Thus $\beta$ serves as a tunable knob that trades off geometric adherence against numerical sensitivity.

\paragraph{Temporal flexibility (Case A3).}

\begin{figure}[t]
  \centering
  \includegraphics[width=\textwidth]{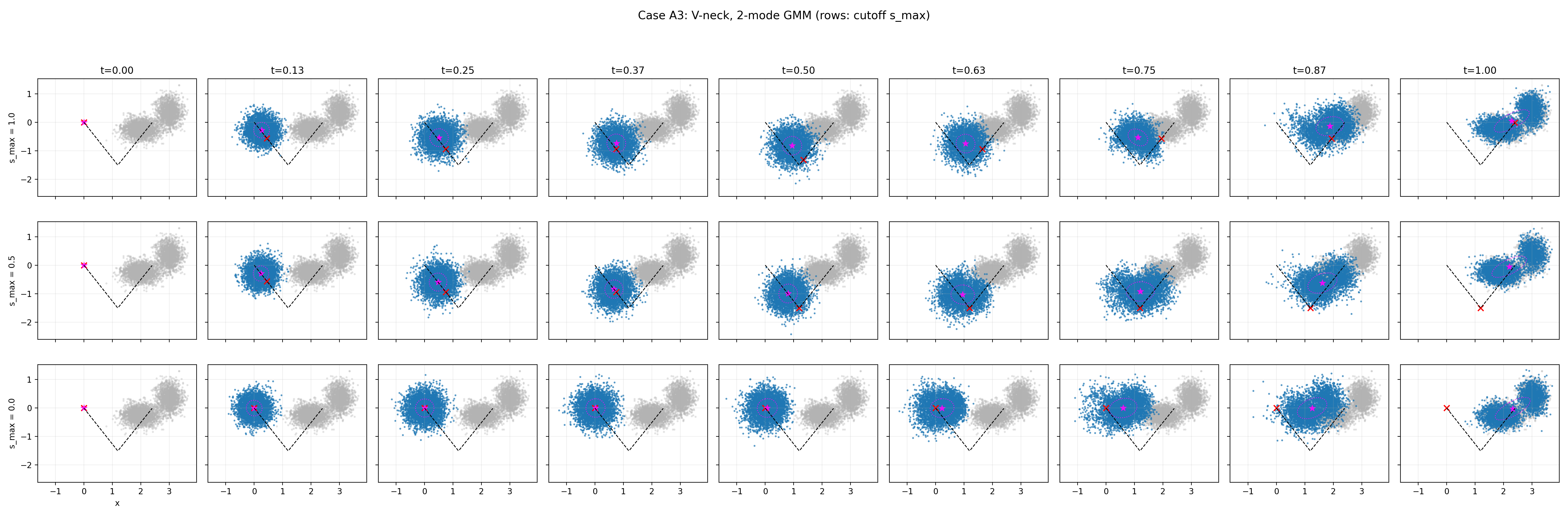}
  \includegraphics[width=\textwidth]{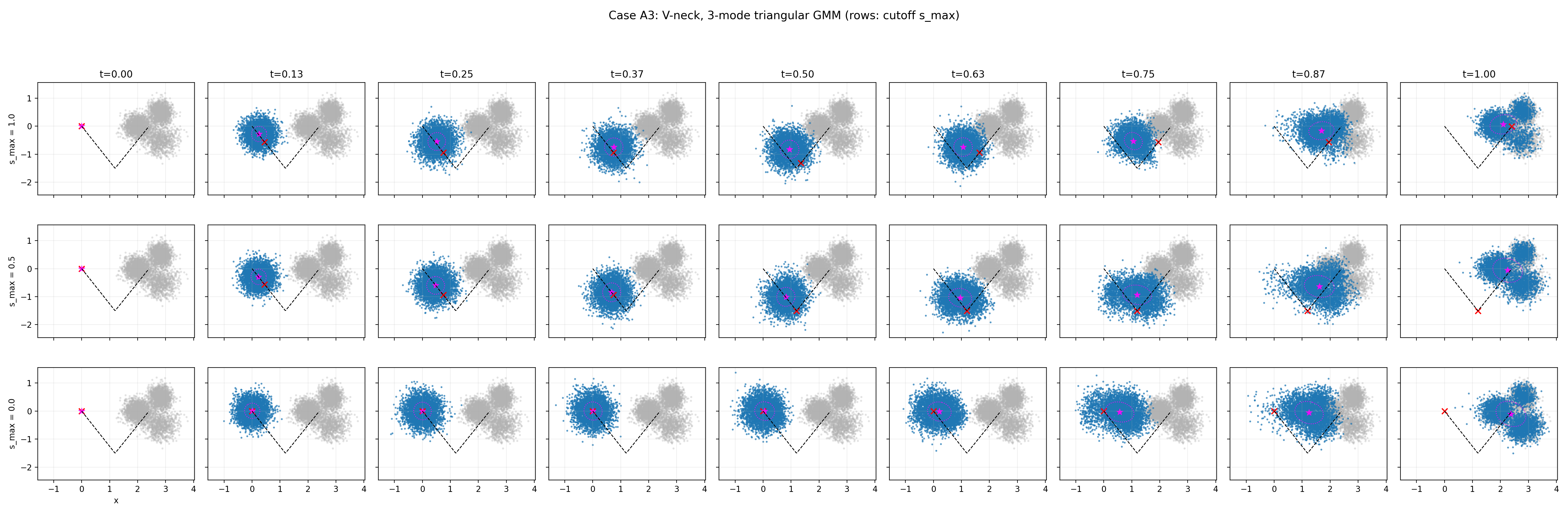}
  \caption{Case~A3: temporal flexibility along a V--neck guide. Each row corresponds to a different cutoff fraction $s_{\max}\in\{1.0,0.5,0.0\}$ in $\nu^{(\mathrm{V},s_{\max})}_t=\nu^{(\mathrm{V})}_{\min(t,s_{\max})}$ for 2 mode (top) and 3 mode (bottom) examples. For $s_{\max}=1$ the guide traverses the full V--neck; smaller $s_{\max}$ produce guides that stop early and remain fixed, forcing GH--PID to complete the transport.  The evolution of the cloud shows how temporal scheduling of the guide interacts with the stochastic dynamics while maintaining correct terminal sampling.} \label{fig:caseA-A3}
\end{figure}

Finally, Fig.~\ref{fig:caseA-A3} illustrates temporal scheduling along a fixed V--neck geometry.  We fix the V--centerline and a constant stiffness $\beta_t\doteq \beta_{\mathrm{V}}$ and vary the cutoff fraction $s_{\max}\in\{1.0,0.8,0.6,0.4,0.2\}$ in the truncated guide $\nu^{(\mathrm{V},s_{\max})}_t=\nu^{(\mathrm{V})}_{\min(t,s_{\max})}$.  Each row corresponds to one $s_{\max}$.

When $s_{\max}=1$ the guide traverses the entire V--neck, and the cloud follows it closely.  As $s_{\max}$ decreases, the guide stops earlier and remains fixed, while the cloud continues to diffuse and drift under GH--PID. The resulting ensembles exhibit progressively greater reliance on intrinsic GH--PID dynamics to reach the terminal region, with corresponding changes in the dispersion and alignment of the cloud.  This experiment emphasizes that, even with fixed geometry and stiffness, re-timing the guide offers a powerful lever for shaping path ensembles without altering the hard terminal constraint.

\paragraph{A3: Partial traversal and temporal allocation.} Fig.~\ref{fig:caseA-A3} explores a different degree of freedom: all schedules share the same straight centerline and constant $\beta$,   but differ in how much of the path is actually traversed by the guide by time $t=1$.  Specifically, we use truncated centerlines $\nu_t^{(\rho)}$ that reach only a fraction $\rho\in\{1.0,0.5,0.0\}$ of the total displacement.

Two phenomena stand out:
\begin{enumerate}
\item \textbf{Cloud retention near origin for small $\rho$.} When the guide does not travel the full length of the corridor, the cloud remains substantially closer to the entrance region throughout the evolution.  The empirical mean shifts proportionally to $\rho$, and the multi-modal splitting occurs later.

\item \textbf{Controlled bias toward the mode nearest the truncated guide.} When the guide stops early, the drift term $b_t^{(-)}(\hat y(t;x) - \nu_t)$ pulls the cloud preferentially toward the mixture component whose mean lies closest to the truncated final guide position.   The resulting mode weights at $t=1$ exhibit a smooth, monotone dependence on $\rho$, illustrating that path length is an effective and interpretable control parameter in GH--PID navigation.
\end{enumerate}

This experiment highlights that geometry and timing of $\nu_t$—even with fixed stiffness—provide fine-grained control over mode selection and mass allocation.

Together, Figs.~\ref{fig:caseA_geometry}--\ref{fig:caseA-A3} present a compact but expressive synopsis of Case~A: GH--PID can encode geometric, confinement, and temporal aspects of navigation directly through the analytic protocol $\Gamma_t=(\nu_t,\beta_t)$, while the stochastic optimal control backbone guarantees that all such protocols sample the same terminal law $p^{(\mathrm{tar})}$.

\subsection{Case B: Results of Protocol Learning}
\label{subsec:caseBresults}

\begin{figure}[t]
  \centering
  \includegraphics[width=0.48\textwidth]{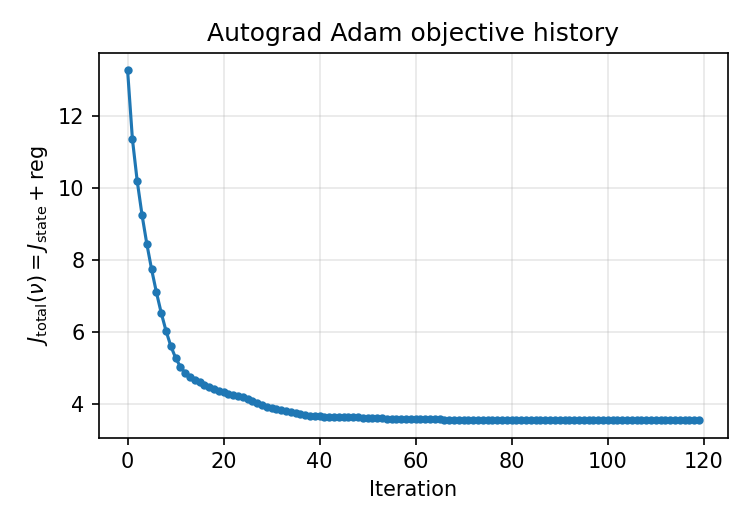}
  \includegraphics[width=0.48\textwidth]{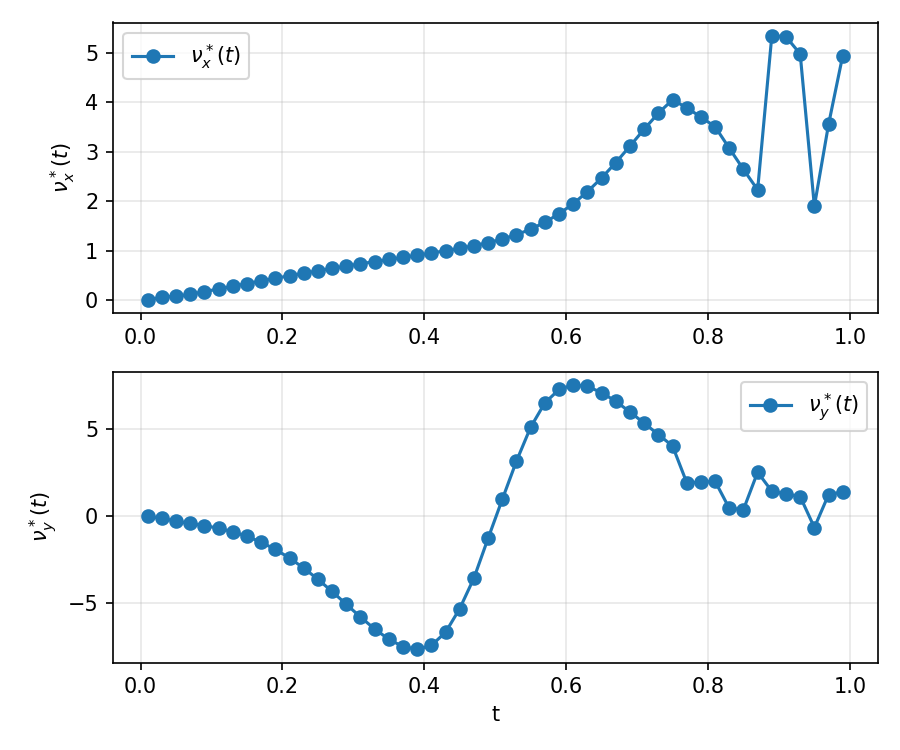}
  \caption{\textbf{Case~B autograd protocol learning.}
  \emph{Top:} Objective history $J(\nu)$ from~\eqref{eq:caseBobjective} over Adam iterations, starting from the teacher/desiderata protocol.  The total cost decreases from $J(\bar\nu)\approx 13.3$ to a best value of about $3.55$, with adaptive step--size reductions when progress stalls.
  \emph{Bottom:} Optimized guidance components $\nu_x^*(t_k)$ and $\nu_y^*(t_k)$ plotted against the PWC midpoints $t_k^{\rm mid}$.  The learned protocol retains a monotone progression along the corridor axis and develops a smooth S--shaped transverse motion, closely aligned with the teacher/desiderata centerline but slightly adjusted by the optimization.}
  \label{fig:caseB_objective_centerline}
\end{figure}

\paragraph{Convergence of the autograd objective.} Fig.~\ref{fig:caseB_objective_centerline} (left) shows the evolution of the total objective $J(\nu)$ in~\eqref{eq:caseBobjective} over autograd--Adam iterations.  Starting from the teacher/desiderata protocol, the objective decreases from
$J(\bar\nu)\approx 13.3$ to a best value of about $3.55$ in roughly $120$ iterations, with the adaptive learning rate flattening residual oscillations once the optimizer enters a narrow basin.  The decay is dominated by the teacher/desiderata term~\eqref{eq:caseB_desiderata_cost}, indicating that relatively modest adjustments of the centerline can substantially improve adherence of the GH--PID trajectories to the desired S--shaped tube while maintaining high terminal likelihood and regularity.

\paragraph{Optimized centerline.} The right panel of Fig.~\ref{fig:caseB_objective_centerline} plots the components of the optimized protocol $\nu^*(t)$ evaluated at the PWC midpoints $t_k^{\rm mid}$.  The $x$--component $\nu_x^*(t)$ progresses monotonically from $x_{\rm in}$ to $x_{\rm out}$, closely tracking the corridor axis, while the $y$--component $\nu_y^*(t)$ develops a smooth, saturating S--shape.  Compared to the straight--axis baseline, the learned guide bends into and out of the two terminal modes in a controlled fashion, reproducing the qualitative shape of the teacher/desiderata curve but with slight adjustments that balance path adherence against terminal cross--entropy and regularization.

\begin{figure}[t]
  \centering
  \includegraphics[width=\textwidth]{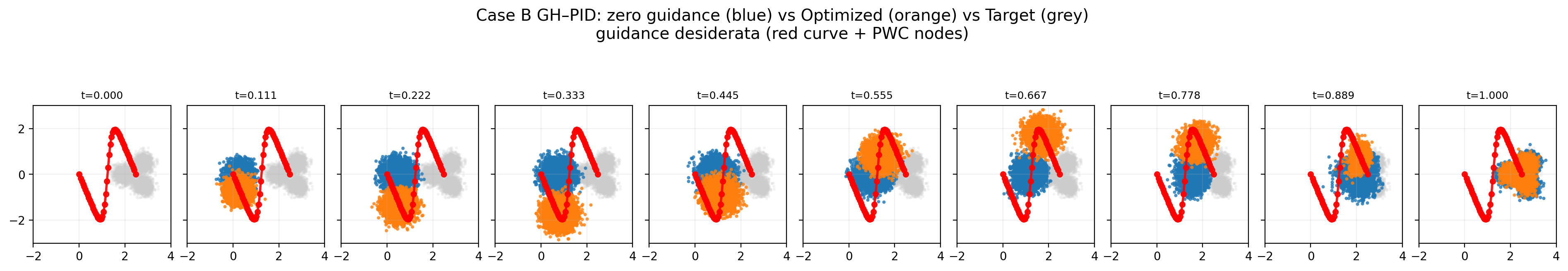}
  \caption{\textbf{Case~B GH--PID snapshots under baseline and optimized guidance.}
  Grey dots: samples from the two--mode GMM target.  Blue dots: GH--PID cloud driven by a straight--axis baseline guide.  Orange dots: GH--PID cloud under the optimized protocol $\nu^*(t)$.  The red curve and nodes show the continuous teacher/desiderata S--shaped centerline and its PWC midpoints.  Columns correspond to increasing diffusion times $t\in[0,1]$.  The optimized protocol steers trajectories into the S--shaped corridor earlier and maintains tighter adherence to the guided tube, while preserving correct terminal sampling of the target GMM.}
  \label{fig:caseB_snapshots}
\end{figure}

\paragraph{Path ensembles: baseline vs optimized guidance.} Fig.~\ref{fig:caseB_snapshots} compares GH--PID path ensembles under three protocols: the target GMM (grey dots), a straight--axis baseline guide (blue), and the optimized guide $\nu^*$ (orange), with the continuous teacher/desiderata curve and its PWC samples overlaid in red.  Each panel shows a snapshot at an increasing diffusion time $t\in[0,1]$, using a fixed spatial window that covers the entire corridor and the two terminal modes.

The baseline protocol transports mass along the corridor axis with relatively weak transverse structure: the cloud remains broad and lags behind the teacher/desiderata centerline in regions of high curvature, and modal splitting occurs relatively late.  In contrast, the optimized protocol bends the cloud into the S--shaped tube earlier in time, with trajectories hugging the guided corridor and separating into the two terminal modes in a controlled way.  Throughout the evolution, the optimized ensemble stays closer to the teacher/desiderata tube, while the terminal empirical distribution remains faithful to the prescribed GMM target.  These experiments demonstrate that even in this simple two--dimensional setting, protocol learning over $\nu_t$ alone can meaningfully reshape path geometry without sacrificing exact terminal matching.

\subsection{Case C: Consensus Navigation from Competing Experts}
\label{subsec:caseCresults}

Case~C implements the multi–expert fusion framework introduced in Section~\ref{subsec:caseC}, where two experts supply both a terminal belief $p{(\tar)}_m$ and a trajectory-level preference $\bar\nu^{(m)}_t$.  According to Eq.~(\ref{eq:three-prod-GMM}) commander assigns \emph{integer trust weights} 
\[
(k_1,k_2)\in\{(1,2),\ (1,1),\ (2,1)\},
\]
and fuses the terminal beliefs via the \emph{exact} product-of-experts rule
\[
p^{(\tar)}(x)\;\propto\;
   \bigl(p_1^{(\tar)}(x)\bigr)^{k_1}\,
   \bigl(p_2^{(\tar)}(x)\bigr)^{k_2}.
\]
Because products of Gaussians remain Gaussians, this fusion yields another GMM whose component means, covariances, and weights are computed in closed form. No approximation is used: the fused law is an \emph{exact} GMM and therefore fully compatible with GH--PID’s analytic backward/forward messages and the closed-form drift $u^*_t$.

At the trajectory level, the same trust weights govern the joint objective -- regularized version of Eq.~(\ref{eq:J-multi})
\begin{align}\label{eq:J-multi-reg}
\mathcal{J}^{\rm (multi-reg)}(\Gamma)
  & = k_1\!\int_0^1 \mathbb{E}[V^{(1)}_t(x_t)]\,dt
  + k_2\!\int_0^1 \mathbb{E}[V^{(2)}_t(x_t)]\,dt\\  \label{eq:joint-obj}
  & + \lambda_{\rm CE}\,J_{\rm CE}(\Gamma;p^{(\tar)})
  + \lambda_{\rm smooth}\!\int_0^1\|\nu''(t)\|^2 dt
  + \lambda_{\rm drift}\!\int_0^1\mathbb{E}\big[\|u^*_t\|^2\big] dt.
\end{align}
The additional penalties promote geometric smoothness and limit excessive drift energy; both are crucial for stable optimization in the multi-expert setting, where conflicting teacher geometries can otherwise produce sharp, oscillatory centerlines.  Only the piecewise-constant centerline $\nu_t$ is optimized; the stiffness profile $\beta_t$ is fixed to the Case~B value.

\begin{figure}[t]
  \centering
  \includegraphics[width=\textwidth]{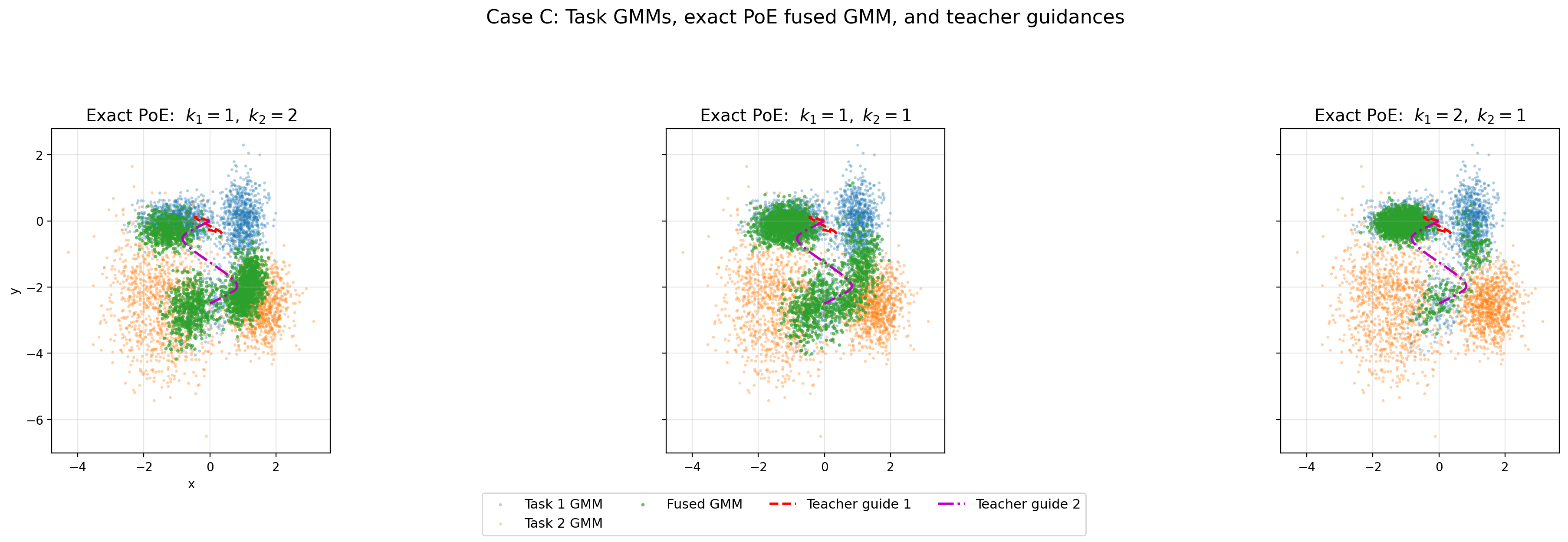}
  \caption{Task GMMs, exact product-of-experts fused GMMs, and teacher-desiderata centerlines for the three trust assignments $(k_1,k_2)\in\{(1,2),(1,1),(2,1)\}$.  The fused distributions change significantly as credibility shifts between the two experts.}
  \label{fig:caseC_gmms_and_guides}
\end{figure}

\paragraph{Fused terminal laws and competing teacher guides.} Fig.~\ref{fig:caseC_gmms_and_guides} shows, for each trust pair $(k_1,k_2)$, the two expert GMMs, the exact PoE fused GMM, and the two teacher curves $\bar\nu^{(1)}_t$ and $\bar\nu^{(2)}_t$.  Despite the experts providing distinct multi-modal GMMs with different endpoints and different geometric semantics, the three fused laws differ substantially: $(1,2)$ strongly favors Expert~2; $(2,1)$ favors Expert~1; and $(1,1)$ yields a symmetric compromise.  These fused GMMs are the exact terminal densities enforced by GH--PID during optimization and sampling.

\begin{figure}[t]
  \centering
  \includegraphics[width=\textwidth]{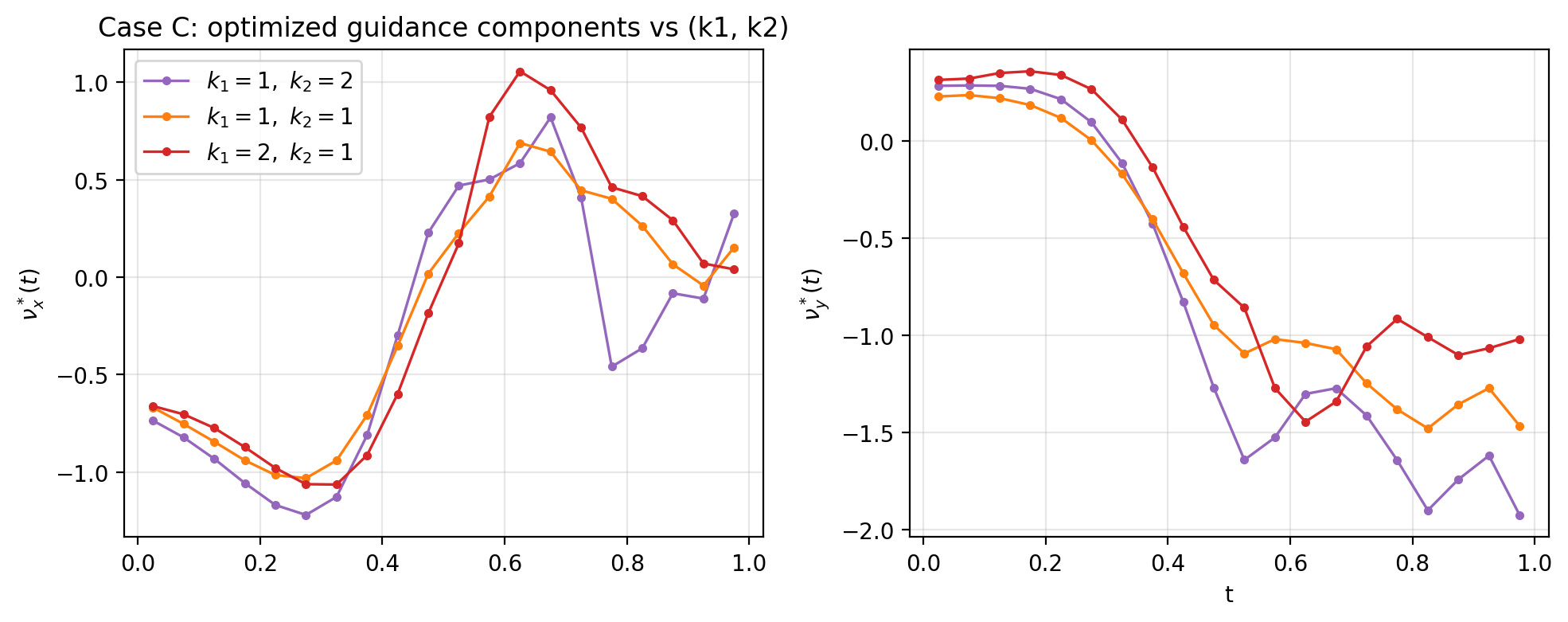}
  \caption{Optimized PWC centerlines $\nu^{*}(t)$ for the three trust configurations. The geometry shifts systematically in response to credibility weights, yielding three distinct consensus trajectories.}
  \label{fig:caseC_nu_vs_t}
\end{figure}

\paragraph{Learned consensus centerlines.} For each trust configuration we run GH--PID with automatic differentiation to obtain the optimal piecewise-constant centerline $\nu^*(t)$.   Fig.~\ref{fig:caseC_nu_vs_t} compares the learned $\nu^*(t)$ against the two teacher/expert curves.  When Expert~2 is more trusted $(k_1,k_2)=(1,2)$, the protocol is visibly pulled toward Expert~2’s endpoint and geometry; the opposite holds for $(2,1)$; and the symmetric case $(1,1)$ produces a clean geometric compromise. The additional smoothness and drift-energy regularizers eliminate the noisy oscillations observed in earlier experiments, producing stable, interpretable consensus centerlines.

\begin{figure}[t]
  \centering
  \includegraphics[width=\textwidth]{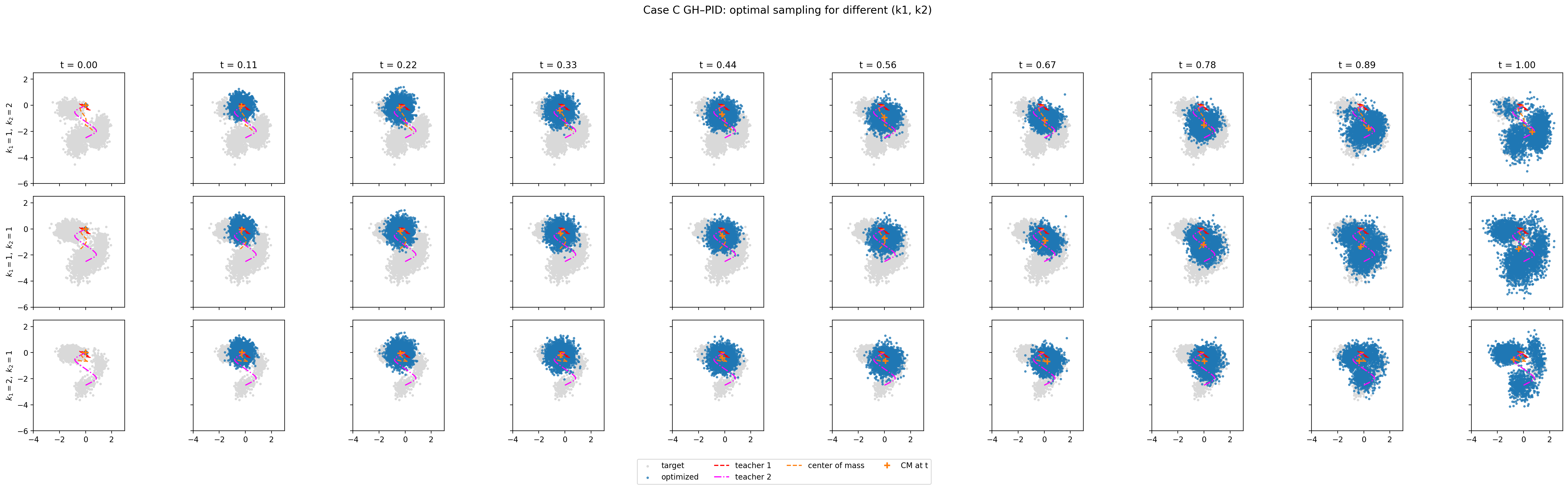}
  \caption{GH--PID evolution under three learned consensus protocols. Rows correspond to trust configurations $(1,2)$, $(1,1)$, and $(2,1)$. Columns show ten uniformly spaced times.  Blue: GH--PID samples.  Gray: fused target samples.  Orange dashed curve: center of mass trajectory, and orange “+”: instantaneous center of mass. The learned protocol steers the ensemble t the appropriate fused terminal law while adopting a globally negotiated geometry.}
  \label{fig:caseC_snapshots_multi_trust}
\end{figure}

\paragraph{Time-resolved GH--PID sampling.} Fig.~\ref{fig:caseC_snapshots_multi_trust} displays ten time frames for each trust configuration.  Each row corresponds to a different $(k_1,k_2)$;  columns progress from $t\approx 0$ to $t\approx1$. Blue points show GH--PID samples; gray points show samples from the exact fused target; the orange dashed curve shows the \emph{evolving center of mass} with a green cross marking the instantaneous CM.

Three key findings emerge:

\begin{enumerate}
\item \textbf{Consensus geometry.}  The ensemble follows the learned protocol $\nu^*(t)$ rather than either teacher curve individually.  GH--PID implements the negotiated geometry encoded by the commander’s trust assignment.

\item \textbf{Trust-dependent mode flow.}  Under $(1,2)$ the mass is steered toward Expert~2’s favored terminal region; under $(2,1)$ it flows toward Expert~1’s; and $(1,1)$ produces a balanced splitting and merging consistent with the exact fused GMM.

\item \textbf{Exact terminal matching.}  
      In all three cases the final-time ensemble matches the fused GMM with high fidelity, confirming that GH--PID remains robust and accurate even under conflicting multi-expert guidance.
\end{enumerate}

\vspace{1ex}
\paragraph{Summary.} Case~C demonstrates that GH--PID acts as a principled \emph{consensus engine}: the same trust weights $(k_1,k_2)$ simultaneously fuse the experts’ terminal beliefs and their trajectory preferences, while the learned protocol $\nu^*(t)$ provides an interpretable geometric compromise. Because GH--PID maintains exact terminal sampling and analytic tractability, it is well suited for multi-expert navigation tasks involving heterogeneous planners, sensors, or AI demonstrators.

\section{Discussion and Outlook}\label{sec:discussion}

This work developed a fully analytic, guidance–centric formulation of 
Guided Harmonic Path-Integral Diffusion (GH--PID) for navigation and 
stochastic optimal transport (SOT) tasks under hard terminal constraints. 
The central technical contribution is a unified framework in which 
(i) the terminal constraint is matched \emph{exactly} via the backward message; 
(ii) the drift is obtained in closed form as a smoothed, instantaneous 
optimal controller; and 
(iii) the time-dependent protocol $\Gamma_t = (\beta_t,\nu_t)$ acts as a 
low-dimensional variational ansatz for shaping the geometry of stochastic 
trajectories.  
The diagnostics introduced in this paper---terminal fidelity, centerline 
adherence, ensemble variability, drift effort, and center-of-mass flow---enable 
quantitative evaluation and gradient-based optimization of~$\Gamma_t$.

The three case studies in Sections~4--5 highlight progressively richer 
capabilities of GH--PID:
\begin{enumerate}
\item \textbf{Case A (Hand-crafted protocols).}  
      By isolating the effects of geometry, stiffness, and time 
      reparameterization, we demonstrated how the guide $\nu_t$ controls 
      curvature-induced lag, mode formation, and splitting thresholds, while 
      $\beta_t$ determines confinement, exploration, and numerical stability.  
      These experiments identified the functional degrees of freedom that most 
      strongly influence path semantics.

\item \textbf{Case B (Single-task protocol learning).}  
      With a fixed task-driven terminal GMM, we optimized a piecewise-constant 
      centerline and showed that modest learning over $\nu_t$ can significantly 
      reduce integrated cost while maintaining exact terminal sampling.  
      The combination of analytic GH--PID drift with automatic differentiation 
      enabled stable optimization despite nonconvexity of the landscape cost.

\item \textbf{Case C (Multi-expert fusion).}  
      Two experts provide both terminal beliefs and trajectory-level 
      preferences.  
      Using exact product-of-experts (PoE) fusion with integer exponents 
      $(k_1,k_2)$, we constructed consensus terminal densities that remain 
      exact GMMs.  
      The same trust weights also fused the two trajectory-level objectives.  
      The resulting optimized centerlines $\nu^{*}(t)$ provided interpretable, 
      credibility-aware consensus trajectories that accurately guided mass to 
      the fused terminal distribution.  
      This case established GH--PID as a principled consensus engine for 
      heterogeneous navigational inputs.
\end{enumerate}

Together, these results show that GH--PID is an analytically transparent, 
computationally efficient, and physically interpretable approach to empirical 
SOT with exact terminal constraints.  
The methodology supports a rich interplay between geometry, stochasticity, 
and optimal transport, yielding interpretable guidance signals and fully 
differentiable protocol learning.

\subsection*{Future Directions}

Several natural extensions follow from the analytic structure and experimental 
findings of this work.  
We highlight four particularly promising directions.

\paragraph{(1) Richer optimization over quadratic potentials.}
The harmonic potentials underlying GH--PID permit analytic backward/forward 
messages and closed-form drifts.  
Future work will explore higher-dimensional, anisotropic, and coupled 
quadratic potentials in general position.  
Such potentials support obstacle-aware guidance, nonuniform confinement, 
and richer funnel/tunnel transitions while preserving linear solvability.

\paragraph{(2) Navigation and monitoring by ensembles of agents in 2D/3D.}
A natural application domain is large-scale navigation, monitoring, and 
surveillance by ensembles of autonomous agents.  
GH--PID’s analytic drift is well suited for cooperative exploration, 
uncertainty-aware planning under sensing constraints, multi-exit missions, 
and time-varying objectives.  
We envision incorporating GH--PID as an interpretable and controllable motion 
primitive within real-time autonomy stacks.

\paragraph{(3) Integration of environmental uncertainty and turbulence.}
Building on recent advances in physics-guided reinforcement learning 
(e.g.,~\cite{koh_physics-guided_2025}), an important direction is to incorporate 
stochastic or turbulent environmental fields into the GH--PID drift.  
This includes biased or anisotropic noise models, partially known environmental 
flows, online inference of physical parameters, and robustness to 
worst-case disturbances.  
Such extensions would yield hybrid analytic--data-driven autonomy frameworks.

\paragraph{(4) Mean-field GH--PID and broadcast population guidance.}
Following mean-field control theory (e.g.,~\cite{metivier_mean-field_2020}), 
GH--PID can be extended to settings where 
each agent follows an autonomous GH--PID policy while a central commander 
broadcasts information about the instantaneous population distribution.  
Agents then blend private objectives with global mean-field guidance, enabling 
emergent coordination without centralized micromanagement.

\paragraph{(5) Molecular and chemical design via navigation in energy landscapes.}
Another promising direction is chemical and molecular navigation, where the 
state space consists of physically meaningful molecular degrees of freedom.  
GH--PID is naturally suited for navigating time-dependent energy landscapes, 
targeting complex terminal distributions, guiding barrier-crossing events, 
and integrating chemical intuition into the guidance protocol~$\nu_t$.  
Such problems closely mirror the navigation tasks studied in this paper, with 
energy landscapes replacing geometric corridors.

\paragraph{Outlook.}
The analytic structure of GH--PID, the differentiability of the full 
sampling--optimization pipeline, and the empirical SOT perspective together 
provide a powerful new toolset for problems where geometry, uncertainty, and 
multi-objective preferences interact.  
Future work will develop higher-order potentials, physics-aware multi-agent 
control, and applications in scientific domains where guided navigation through 
complex landscapes is essential.

\section*{Acknowledgments}

The author acknowledge support of start up at the University of Arizona. A substantial portion of this work was carried out while the author was partially funded by a mini-sabbatical Faculty Fellow Program at Lawrence Livermore National Laboratory (LLNL). I gratefully acknowledge the support, scientific engagement, and encouragement of colleagues in the Center for Applied Scientific Computing (CASC) at LLNL.

In the interest of reproducibility, all Python/Jupyter code used to generate the figures and experiments reported in this paper is openly available at \href{https://github.com/mchertkov/GuidedPID}{github.com/mchertkov/GuidedPID}.

The author used large language models (OpenAI ChatGPT) to assist with editing, code refactoring, and organization of computational experiments. All mathematical derivations, scientific claims, and final code were independently checked, validated, and approved by the author.

\printbibliography

\appendix

\section{Green Functions and Riccati Equations for Quadratic Potential}\label{sec:Riccati}

Solutions of Eqs.~(\ref{eq:kfp_backward},\ref{eq:kfp_forward}) with the guided quadratic/harmonic potential \eqref{eq:guided_potential} are 
\begin{align}
\label{eq:Gminus_gauss}
G_t^{(-)}(x\!\mid\!  y) &\;\propto\; 
\exp\!\Biggl(\!
-\frac{a_t^{(-)}}{2}\,\|x-\nu_t\|^2 
+ b_t^{(-)} (x-\nu_t)^\top (y-\nu_t)\\ \nonumber 
& -\frac{c_t^{(-)}}{2}\,\|y-\nu_t\|^2 
+ (r_t^{(-)})^\top (x-\nu_t) + (s_t^{(-)})^\top (y-\nu_t)\!\Biggr),\\
\label{eq:Gplus_gauss}
G_t^{(+)}(y\!\mid\!  0) &\;\propto\;
\exp\!\left(\!
-\frac{a_t^{(+)}}{2}\,\|y-\nu_t\|^2 + (s_t^{(+)})^\top (y-\nu_t)\!\right).
\end{align}
The quadratic coefficients obey the Riccati system
\begin{equation}
\label{eq:riccati_quad}
\mp\,\dot a_t^{(\pm)} + \beta_t = \big(a_t^{(\pm)}\big)^2,\qquad 
\dot b_t^{(-)} = a_t^{(-)} b_t^{(-)},\qquad 
\dot c_t^{(-)} = \big(b_t^{(-)}\big)^2,
\end{equation}
and the linear terms satisfy
\begin{equation}
\label{eq:riccati_lin}
\dot r_t^{(-)} + \big(a_t^{(-)} - b_t^{(-)}\big)\,\dot\nu_t = a_t^{(-)} r_t^{(-)},\ 
\dot s_t^{(-)} + \big(c_t^{(-)} - b_t^{(-)}\big)\,\dot\nu_t = -\, b_t^{(-)} r_t^{(-)},\ 
\dot s_t^{(+)} + a_t^{(+)} \dot\nu_t = -\, a_t^{(+)} s_t^{(+)}.
\end{equation}
Boundary behavior is
\begin{equation}
\label{eq:bc_limits}
t\to 1^-:\;\; a_t^{(-)},b_t^{(-)},c_t^{(-)} \sim \frac{1}{1-t},\;\; r_t^{(-)}\to 0,\;\; s_t^{(-)}\to 0;
\
t\to 0^+:\;\; a_t^{(+)} \sim \frac{1}{t},\;\; s_t^{(+)}\to 0.
\end{equation}
\emph{Remark.} Each instantaneous generator $\tfrac{1}{2}\Delta - V_t$ is self-adjoint on $L^2(\mathbb{R}^d)$, but the time-ordered family need not commute; therefore $G_t^{(-)}$ is not symmetric in $(x,y)$ for general time-varying $\beta_t,\nu_t$.

\paragraph{Reweighting Gaussian and its mean.}
Define
\begin{equation}
\label{eq:K_def}
K_t \;\doteq\; c_t^{(-)} - a_1^{(+)},
\qquad
\psi_t \;\doteq\; s_t^{(-)} - s_1^{(+)}.
\end{equation}
Then the $y$-dependence of $\Delta$ in \eqref{eq:Delta_def} is quadratic, and the \emph{reweighting} density
$\rho(y\!\mid\!  t;x;\Gamma)\propto \exp\!\big(-\Delta(t;x;y;\Gamma)\big)$ is Gaussian with covariance $(1/K_t)\,I_d$ and mean
\begin{equation}
\label{eq:mu_rho}
\mu_t(x) \;=\; \nu_t \;+\; \frac{b_t^{(-)}}{K_t}\,\big(x-\nu_t\big) \;+\; \frac{\psi_t}{K_t}\,.
\end{equation}
Equivalently,
\begin{equation}
\label{eq:rho_density}
\rho(y\!\mid\!  t;x;\Gamma) \;=\; \mathcal{N}\!\left(y \Big| \mu_t(x),\; \frac{1}{K_t}\,I_d\right).
\end{equation}

\paragraph{Gradient of the backward kernel.}
From \eqref{eq:Gminus_gauss},
\begin{equation}
\label{eq:grad_logGminus}
\nabla_x \log G_t^{(-)}(x\!\mid\!  y) \;=\; -\,a_t^{(-)}\big(x-\nu_t\big) \;+\; b_t^{(-)}\big(y-\nu_t\big) \;+\; r_t^{(-)}.
\end{equation}
We will use \eqref{eq:grad_logGminus} together with \eqref{eq:rho_density} to obtain compact expressions for the optimal drift in Appendix~\ref{sec:yhat}.

\subsection{PWC formulas for the moving guide}\label{sec:PWC-guide}

\subsubsection*{Why PWC?}

Using a piece–wise–constant (PWC) protocol $\Gamma_t=\{\beta_t,\nu_t\}$ yields fully analytic updates for the Riccati and linear coefficients of the Gaussian Green functions (cf. Appendix~A), thus avoiding stiffness and accumulated quadrature error when numerically integrating the Riccati ODEs. In practice this gives stable, closed–form “in–piece” evolution and explicit interface \emph{kicks} when the guide center $\nu_t$ jumps at partition times.

\noindent\textbf{Setup.}
Split $[0,1]$ into $K$ equal pieces and make $\Gamma_t=(\beta_t,\nu_t)$ PWC:
\begin{equation*}
k=1,\dots,K,\quad  t\in\bigl[(k{-}1)/K,\,k/K\bigr]:\qquad \beta_t=\beta_k,\quad \nu_t=\nu_k.
\end{equation*}
Throughout, $(a_t^{(\pm)},b_t^{(-)},c_t^{(-)})$ denote the quadratic coefficients of the
backward/forward kernels, while $(r_t^{(-)},s_t^{(-)},s_t^{(+)})$ are the corresponding linear terms (see Eqs.~(\ref{eq:Gminus_gauss}-\ref{eq:bc_limits})).

\subsubsection{Quadratic coefficients for PWC $\beta$}\label{sec:quadratic-PWC}

Assuming continuity in time and using the asymptotics from Appendix~A, the \emph{backward} branch on the last piece gives
\begin{align}
\label{eq:PWC-K}
t\in[1{-}1/K,\,1]:\quad
a_t^{(-)}=c_t^{(-)}\!=\!\sqrt{\beta_K}\,\coth\bigl((1{-}t)\sqrt{\beta_K}\bigr),\ b_t^{(-)}\!=\!\frac{\sqrt{\beta_K}}{\sinh\bigl((1{-}t)\sqrt{\beta_K}\bigr)}.
\end{align}
Then for $k=K{-}1,\dots,1$ and $t\in\bigl[(k{-}1)/K,\,k/K\bigr]$,
\begin{align}
\nonumber 
&a_t^{(-)}=\sqrt{\beta_k}\,
\frac{a_{k/K}^{(-)}\!+\!\sqrt{\beta_k}\,\tanh\left(\sqrt{\beta_k}(k/K{-}t)\right)}
{\sqrt{\beta_k}\!+\! a_{k/K}^{(-)}\,\tanh\left(\sqrt{\beta_k}(k/K{-}t)\right)},\ 
b_t^{(-)}
=b_{k/K}^{(-)}\,
\sqrt{\frac{\beta_k-\bigl(a_t^{(-)}\bigr)^2}{\beta_k-\bigl(a_{k/K}^{(-)}\bigr)^2}},\\ \label{eq:PWC-k}
&c_t^{(-)}=c_{k/K}^{(-)}+\frac{\bigl(b_{k/K}^{(-)}\bigr)^2}{\beta_k-\bigl(a_{k/K}^{(-)}\bigr)^2}\,
\bigl(a_{k/K}^{(-)}-a_t^{(-)}\bigr).
\end{align}
For the \emph{forward} branch,
\begin{align}
\label{eq:PWC-K+}
t\in[0,\,1/K]:\qquad a_t^{(+)}=\sqrt{\beta_1}\,\coth\!\bigl(t\sqrt{\beta_1}\bigr),
\end{align}
and for $t\in\bigl[k/K,\,(k{+}1)/K\bigr],\ k=1,\dots,K{-}1$
\begin{align}
\label{eq:PWC-k+}
a_t^{(+)}\!=\!\sqrt{\beta_{k+1}}\,
\frac{1+\rho_{k+1}(t)}{1-\rho_{k+1}(t)},\ \rho_{k+1}(t)\!=\!\exp\bigl(-2\sqrt{\beta_{k+1}}(t{-}k/K)\bigr)\,
\frac{a_{k/K}^{(+)}-\sqrt{\beta_{k+1}}}{a_{k/K}^{(+)}+\sqrt{\beta_{k+1}}}.
\end{align}

\emph{Derivation remark.} For the ``$-$'' branch $\dot a=a^2-\beta$, the M\"{o}bius variable $u=(a-\sqrt\beta)/(a+\sqrt\beta)$ solves $\dot u=+2\sqrt\beta\,u$, giving the $\tanh$ form above. For the ``$+$'' branch $\dot a=\beta-a^2$, the same $u$ obeys $\dot u=-2\sqrt\beta\,u$, yielding the exponential M\"{o}bius map.

\subsubsection{Linear coefficients with PWC $\nu$}\label{sec:linear-PWC}
Inside each piece $\nu$ is constant, hence $\dot\nu_t=0$ and the linear ODEs (\ref{eq:riccati_lin}) reduce to
\[
\dot r_t^{(-)}=a_t^{(-)}\,r_t^{(-)},\qquad
\dot s_t^{(-)}=-\,b_t^{(-)}\,r_t^{(-)},\qquad
\dot s_t^{(+)}=-\,a_t^{(+)}\,s_t^{(+)},
\]
and in transition between pieces there will be kicks (jumps) associated with jumps in $\nu$.

Therefore in the reverse branch we initialize $s_{1^-}^{(-)}=r_{1^-}^{(-)}=0$; assume that $a_t^{(-)}, b_{t}^{(-)}, c_{t}^{(-)}$ are already computed (in Section \ref{sec:quadratic-PWC}) and proceed with $k=K,\dots,1$:
\begin{align}
\label{eq:PWC-guide-minus-inpiece}
t\in \left[\frac{(k-1)^+}{K},\,\frac{k^-}{K}\right]: &\quad 
r_t^{(-)}=r_{k^-/K}^{(-)}\exp\left(\int_{k^-/K}^t a^{(-)}_\tau d\tau\right)
=r_{k^-/K}^{(-)}\;\frac{b_t^{(-)}}{b_{k/K}^{(-)}},\\ \nonumber & \quad 
s_t^{(-)}=s_{k^-/K}^{(-)}-\int_{k^-/K}^t b^{(-)}_\tau r^{(-)}_\tau d\tau
=s_{k^-/K}^{(-)}\;+\;\frac{r_{k^-/K}^{(-)}}{\,b_{k/K}^{(-)}\,}\,\Bigl(c_t^{(-)}-c_{k^-/K}^{(-)}\Bigr),\\ 
\label{eq:PWC-guide-minus-jump}
t=\frac{(k-1)^-}{K}:&\quad  r_{(k-1)^-/K}^{(-)} =r_{(k-1)^+/K}^{(-)}-\bigl(a_{(k-1)/K}^{(-)}-b_{(k-1)/K}^{(-)}\bigr)\left(\nu_{k-1}-\nu_{k}\right),\\ \nonumber &\quad  s^{(-)}_{(k-1)^-/K}\;=\;
s^{(-)}_{(k-1)^+/K}\;-\;\bigl(c^{(-)}_{(k-1)/K} - b^{(-)}_{(k-1)/K}\bigr)\,\bigl(\nu_{k-1} - \nu_k\bigr)\,.
\end{align}
Here, in Eq.~(\ref{eq:PWC-guide-minus-inpiece}), we accounted for the ODEs governing dynamics of $a_t^{(-)}, b_{t}^{(-)}, c_{t}^{(-)}$; and Eq.~(\ref{eq:PWC-guide-minus-jump}) describe discontinuity (jump) of $r_t$ and $s_t$  at $t=(k-1)/K$, and $(k-1)^{\pm}/K$ is introduced to index values (of $r_t$ and $s_t$) at $\pm$ sides of the jump.

Now evaluating $s^{(+)}_t$ forward in time we initialize it with $s^{(+)}_{0^+/K}$ and then 
march forward with $k=1,\cdots,K$  thus arriving at
\begin{eqnarray}\nonumber 
t\in \left[\frac{(k-1)^+}{K},\frac{k^-}{K}\right]:& \quad  s_t^{(+)} & =s_{(k-1)^+/K}^{(+)}\exp \left(-\int_{(k-1)/K}^{t} a_\tau^{(+)}\,d\tau\right)\\ \label{eq:st+} & & = s_{(k-1)^+/K}^{(+)} 
e^{-\sqrt{\beta_k}\,\left(t-\frac{k-1}{K}\right)}\,
\frac{a^{(+)}_t+\sqrt{\beta_k}}{a^{(+)}_{(k-1)/K}+\sqrt{\beta_k}},\\
.\label{eq:PWC-guide-plus-jump}
t=\frac{k^+}{K}:&\quad  s_{k^+/K}^{(+)} & =s_{k^-/K}^{(+)}-a_{k/K}^{(+)}\left(\nu_{k+1}-\nu_k\right),
\end{eqnarray}
where we resolve the integral in Eq.~(\ref{eq:st+}) integrating Eq.~(\ref{eq:PWC-k+}) over time, and Eq.~(\ref{eq:PWC-guide-plus-jump}) represents jumps in $s_t^{(+)}$.

\subsubsection*{Algorithmic summary}

\begin{enumerate}
\item Backward pass: compute $(a^{(-)},b^{(-)},c^{(-)})$ via \eqref{eq:PWC-K}–\eqref{eq:PWC-k}; set $r^{(-)}{=}s^{(-)}{=}0$ on $[1{-}1/K,1]$; for $k=K{-}1,\dots,1$ apply \eqref{eq:PWC-guide-minus-jump} at $t_k$ and propagate on $[(k{-}1)/K,\,k/K]$ using \eqref{eq:PWC-guide-minus-inpiece}.

\item Forward pass: compute $a^{(+)}$ via \eqref{eq:PWC-K+}–\eqref{eq:PWC-k+}; set $s^{(+)}= 0$ on $[0,1/K]$; for $k=1,\dots,K{-}1$ apply \eqref{eq:PWC-guide-plus-jump} at $t_k$ and propagate inside $[k/K,(k{+}1)/K]$ according to Eq.~(\ref{eq:st+}). 
\end{enumerate}

\subsubsection*{General Remarks}
(i) All \emph{quadratic} coefficients are continuous across $t_k$; only the \emph{linear} terms receive kicks proportional to $\Delta\nu$. (ii) On any piece with constant $\beta_k$, $J_t=(a_t^{(-)})^2-(b_t^{(-)})^2\doteq \beta_k$, so Eq.~\eqref{eq:PWC-guide-minus-inpiece} can also be written as
\[
r_t^{(-)}\,=\,r_{k/K^-}^{(-)}\,
\sqrt{\frac{\beta_k-(a_t^{(-)})^2}{\beta_k-(a_{k/K^-}^{(-)})^2}},\quad  
s_t^{(-)}\,=\,s_{k/K^-}^{(-)}+\frac{r_{k/K^-}^{(-)}}{b_{k/K^-}^{(-)}}\Bigl(c_t^{(-)}-c_{k/K^-}^{(-)}\Bigr).
\]

\section{Analytic $\hat y(t;x)$ for Gaussian–Mixture Targets}\label{sec:yhat}

We treat the case where the target is a Gaussian mixture with component–specific full covariances,
\begin{equation}\label{eq:G-Mix-full}
p^{(\mathrm{tar})}(y)=\sum_{n=1}^N \varrho_n\,\mathcal{N}\!\big(y;\mu_n,\Sigma_n\big),
\qquad
\varrho_n\ge 0,\ \sum_{n=1}^N \varrho_n=1,
\end{equation}
with $\Sigma_n\in\mathbb{R}^{d\times d}$ symmetric positive–definite. The probe density defined in
\eqref{eq:probe_density} can be written as a reweighted target,
\[
p(y\mid t;x;\Gamma)
=\frac{p^{(\mathrm{tar})}(y)\,\rho(y\mid t;x;\Gamma)}
       {\int p^{(\mathrm{tar})}(y')\,\rho(y'\mid t;x;\Gamma)\,dy'}.
\]
For GH–PID the reweighting factor is a (time–local) Gaussian,
\begin{equation}\label{eq:rho-gauss}
\rho(y\mid t;x;\Gamma)\;\propto\;
\exp\!\Big(-\tfrac12 K_t\,\|y-\mu_t(x)\|_2^2\Big),
\end{equation}
where $K_t>0$ (the effective scalar precision from Appendix~\ref{sec:Riccati}) and $\mu_t(x)$ is the known affine function of $x$ determined by the PWC coefficients (see Appendix~\ref{sec:Riccati}). Multiplying the Gaussian mixture \eqref{eq:G-Mix-full} by \eqref{eq:rho-gauss} and integrating component–wise gives the Gaussian–mixture posterior
\begin{equation}\label{eq:posterior-mix}
p(y\mid t;x;\Gamma)\ \propto\
\sum_{n=1}^N \varrho_n\,w_n(t;x)\,
\mathcal{N}\!\big(y;\,\tilde{\mu}_n(t;x),\,\tilde{\Sigma}_n(t)\big),
\end{equation}
with (for each $n$)
\begin{align}
\label{eq:Gmix-post-cov-full}
\tilde\Sigma_n(t)
&=\Big(\Sigma_n^{-1}+K_t I_d\Big)^{-1}
=\Sigma_n\Big(I_d+K_t\Sigma_n\Big)^{-1},\\[3pt]
\tilde\mu_n(t;x)
&=\tilde\Sigma_n(t)\Big(\Sigma_n^{-1}\mu_n+K_t\,\mu_t(x)\Big)
=\Big(I_d+K_t\Sigma_n\Big)^{-1}\Big(\mu_n+K_t\Sigma_n\,\mu_t(x)\Big),\\[3pt]
\label{eq:Gmix-weight-full}
w_n(t;x)
&=\mathcal{N}\!\Big(\mu_t(x);\ \mu_n,\ \Sigma_n+\tfrac{1}{K_t}I_d\Big)\\
&=\frac{\exp\!\Big(-\tfrac12\big(\mu_t(x)-\mu_n\big)^\top
\big(\Sigma_n+\tfrac{1}{K_t}I_d\big)^{-1}\big(\mu_t(x)-\mu_n\big)\Big)}
{\sqrt{(2\pi)^d\det\!\big(\Sigma_n+\tfrac{1}{K_t}I_d\big)}}.
\nonumber
\end{align}

Therefore, the predicted (posterior at $t=1$) state map is the mixture mean
\begin{equation}\label{eq:hat-x-G-mix-full}
\hat{y}(t;x)
=\frac{\sum_{n=1}^N \varrho_n\,w_n(t;x)\,\tilde\mu_n(t;x)}
       {\sum_{n=1}^N \varrho_n\,w_n(t;x)}.
\end{equation}

\subsection{Stable evaluation (large $d$, ill–conditioned $\Sigma_n$, extreme $K_t$)}
For robustness, avoid explicit matrix inverses and work with Cholesky factors and log–weights.

\paragraph{(a) Mixture weights in log–space.}
Define the convolution covariances
\[
S_n(t)\;\doteq\;\Sigma_n+\tfrac{1}{K_t}I_d,
\qquad S_n=L_n L_n^\top\ \ (\text{Cholesky}).
\]
Compute $s_n$ by two triangular solves
$L_n s_n=\mu_t(x)-\mu_n$. Then
\[
\log w_n(t;x)=-\tfrac12\|s_n\|_2^2\;-\;\sum_{i=1}^d\log (L_n)_{ii}\;-\;\tfrac{d}{2}\log(2\pi).
\]
Use a log-sum-exp to normalize:
\[
\tilde w_n
=\frac{\exp\!\big(\log w_n-\max_j \log w_j\big)}
{\sum_m \exp\!\big(\log w_m-\max_j \log w_j\big)}.
\]
Only ratios of weights enter \eqref{eq:hat-x-G-mix-full}, so the additive constant cancels.

\paragraph{(b) Posterior mean/covariance without $\Sigma_n^{-1}$.}
Use the algebraically equivalent forms (no explicit inverse of $\Sigma_n$):
\begin{align*}
\tilde\Sigma_n(t)&=\Sigma_n\big(I_d+K_t\Sigma_n\big)^{-1},\\
\tilde\mu_n(t;x)&=\big(I_d+K_t\Sigma_n\big)^{-1}\!\Big(\mu_n+K_t\Sigma_n\,\mu_t(x)\Big).
\end{align*}
Factor $A_n(t)\doteq I_d+K_t\Sigma_n=R_n R_n^\top$ (Cholesky). Then
\[
\tilde\mu_n(t;x)
=R_n^{-\top}R_n^{-1}\Big(\mu_n+K_t\Sigma_n\,\mu_t(x)\Big),
\qquad
\tilde\Sigma_n(t)=R_n^{-\top}R_n^{-1}\,\Sigma_n,
\]
obtained via triangular solves only. In practice, compute $\tilde\mu_n$ with two solves; form
$\tilde\Sigma_n$ only if explicitly needed for downstream uncertainty summaries.

\paragraph{(c) Numerical notes.}
(i) When $K_t$ is very large, $S_n=\Sigma_n+K_t^{-1}I$ is well–conditioned; the weights concentrate near the component closest to $\mu_t(x)$.  
(ii) When $K_t$ is small, weight discrimination weakens; use the log–space normalization above.  
(iii) If $\Sigma_n$ vary strongly across $n$, pre–whiten $y$ (one global linear transform) to improve conditioning, apply the formulas in whitened coordinates, and unwhiten $\hat y$ at the end.  
(iv) All steps are batched over $n$ and $t$ for GPU efficiency.

\medskip
\noindent\emph{Outcome.} Equations \eqref{eq:Gmix-post-cov-full}–\eqref{eq:hat-x-G-mix-full} give a
\textbf{fully analytic}, numerically stable $\hat y(t;x)$ for GMM targets with full covariances,
compatible with the PWC GH–PID machinery (Appendix~\ref{sec:Riccati}) and ready for PyTorch implementation using Cholesky solves and log–space weighting.

\section{Optimal Drift}
\label{sec:ustar}
Starting from \eqref{eq:score_from_probe},
\[
u_t^\ast(x;\Gamma)
=\mathbb{E}_{\,y\sim p(\cdot\mid t;x;\Gamma)}
\!\big[-\,\nabla_x \Delta(t;x;y;\Gamma)\big].
\]
Using the definition
\(
\Delta(t;x;y;\Gamma)
=\!-\,\log G_t^{(-)}(x\mid y;\Gamma)
+\log C(t;x;\Gamma)
\)
and the fact that
\(
\nabla_x\log C(t;x;\Gamma)
=\mathbb{E}_{y\sim\rho(\cdot\mid t;x;\Gamma)}
\big[\nabla_x\log G_t^{(-)}(x\mid y;\Gamma)\big]
\)
(cf.\ \eqref{eq:grad_logGminus} and \eqref{eq:mu_rho}),
we obtain the “two–expectations’’ form
\begin{equation}
\label{eq:ustar_twoexp}
u_t^\ast(x;\Gamma)
=\mathbb{E}_{\,y\sim p(\cdot\mid t;x;\Gamma)}
\!\big[\nabla_x \log G_t^{(-)}(x\mid y;\Gamma)\big]
-\mathbb{E}_{\,y\sim \rho(\cdot\mid t;x;\Gamma)}
\!\big[\nabla_x \log G_t^{(-)}(x\mid y;\Gamma)\big].
\end{equation}

For the guided backward Green function $G_t^{(-)}$, Appendix~\ref{sec:Riccati}
gives the affine–in–$x$ score
\begin{equation}
\label{eq:grad-logGminus-affine}
\nabla_x \log G_t^{(-)}(x\mid y;\Gamma)
=-\,a_t^{(-)}\bigl(x-\nu_t\bigr)
+ b_t^{(-)}\bigl(y-\nu_t\bigr)
+ r_t^{(-)},
\end{equation}
where $a_t^{(-)}>0$ and $b_t^{(-)}>0$ are the quadratic and cross coefficients from the backward Riccati system, $\nu_t$ is the guidance path, and $r_t^{(-)}\in\mathbb{R}^d$ is the linear coefficient associated with the guided backward Green function (see Appendix~\ref{sec:Riccati}).

Inserting \eqref{eq:grad-logGminus-affine} into \eqref{eq:ustar_twoexp} and taking expectations with respect to $p$ and $\rho$ gives
\begin{align*}
u_t^\ast(x;\Gamma)
&= \Bigl(-a_t^{(-)}(x-\nu_t)
        + b_t^{(-)}\bigl(\hat y(t;x;\Gamma)-\nu_t\bigr)
        + r_t^{(-)}\Bigr) \\
&\quad
 -\Bigl(-a_t^{(-)}(x-\nu_t)
        + b_t^{(-)}\bigl(\mathbb{E}_{\rho}[y]-\nu_t\bigr)
        + r_t^{(-)}\Bigr),
\end{align*}
so that the $-a_t^{(-)}(x-\nu_t)$ and $r_t^{(-)}$ terms cancel.  Denoting
\begin{equation}
\label{eq:mu_rho_def}
\mu_t(x;\Gamma)\doteq \mathbb{E}_{\,y\sim \rho(\cdot\mid t;x;\Gamma)}[y]
\end{equation}
(cf.\ \eqref{eq:mu_rho}), we arrive at Eq.~(\ref{eq:ustar_compact_forwardref}) which we reproduce here for convenience \[u_t^\ast(x;\Gamma)= b_t^{(-)}\Bigl(\hat y(t;x;\Gamma)-\mu_t(x;\Gamma)\Bigr).\]

Thus the optimal drift is the backward gain $b_t^{(-)}$ times the discrepancy between the \emph{predicted terminal state} $\hat y(t;x;\Gamma)$ (Appendix~\ref{sec:yhat}) and the reweighting mean $\mu_t(x;\Gamma)=\mathbb{E}_{\rho}[y]$ defined by the auxiliary measure $\rho(\cdot\mid t;x;\Gamma)$.  In particular, $K_t$ enters only implicitly through the construction of $\hat y$ and $\mu_t$ via the Riccati machinery of Appendix~\ref{sec:Riccati}; it does not appear explicitly in \eqref{eq:ustar_compact_forwardref}.

For later reference it is also convenient to record the equivalent “expanded’’ form, obtained by substituting the explicit expression for $\mu_t(x;\Gamma)$ in terms of $(a_t^{(-)},b_t^{(-)},r_t^{(-)},\nu_t)$,
\begin{equation}
\label{eq:ustar_expanded_guided}
u_t^\ast(x;\Gamma)
=-\,a_t^{(-)}\bigl(x-\nu_t\bigr)
+ b_t^{(-)}\bigl(\hat y(t;x;\Gamma)-\nu_t\bigr)
+ r_t^{(-)}.
\end{equation}
In the AdaPID limit ($\nu_t=r_t^{(-)}\doteq  0$) of \cite{chertkov_adaptive_2025}, \eqref{eq:ustar_expanded_guided} reduces to $u_t^\ast(x)=b_t^{(-)}\hat y(t;x)-a_t^{(-)}x$.

\section{Navigation Protocol Templates and Implementation Details}
\label{app:navigation}

This appendix elaborates on the geometric setups, protocol construction, and
implementation details for the navigation experiments of
Section~\ref{sec:navigation}.  In all cases the GH--PID sampler is driven by a
piecewise--constant (PWC) protocol
\[
  \Gamma^{(\mathrm{  PWC})}
    = \{(\nu_k,\beta_k)\}_{k=0}^{K-1},
\]
obtained by discretizing a continuous pair $(\nu_t,\beta_t)$ on $[0,1]$.

We work in $d=2$ for clarity, although the construction extends
straightforwardly to $d=3$ and higher dimensions (latter relevant, e.g. to
navigation in ``chemical'' spaces for molecular design).

\subsection{Case A: Continuous Templates and PWC Protocols}
\label{app:caseA}

We first specify the continuous centerline $\nu_t$ and stiffness profile
$\beta_t$ that encode the corridor geometry and confinement level.  These are
functions
\[
  \nu:[0,1]\to\mathbb{R}^2,\qquad
  \beta:[0,1]\to\mathbb{R}_+,
\]
whose images define a \emph{soft tube} in space--time.  The GH--PID sampler
interacts only with the PWC approximation $\Gamma^{(\mathrm{  PWC})}$; however,
all geometric intuition and visualization are phrased in terms of the
continuous pair $\Gamma=(\nu,\beta)$.

\paragraph{Intrinsic frame and generic centerline.}

Let $x_{\mathrm{in}}=0\in\mathbb{R}^2$ denote the entry point and
$x_{\mathrm{out}}\in\mathbb{R}^2$ the center of mass of the two--mode GMM
target.  We define the axis direction, its unit vector, and a unit normal by
\[
  v = x_{\mathrm{out}} - x_{\mathrm{in}},\qquad
  e = \frac{v}{\|v\|},\qquad
  n = (-e_2,e_1),
\]
and use a scalar parameter $s\in[0,1]$ to measure progress along the axis via
$x_{\mathrm{axis}}(s)=x_{\mathrm{in}}+s\,v$.  All centerlines considered in
Case~A are of the form
\begin{equation}
  \nu(s) = x_{\mathrm{axis}}(s) + \Delta(s)\,n,\qquad s\in[0,1],
  \label{eq:app_centerline_generic}
\end{equation}
for suitable scalar offsets $\Delta(s)$.  In the simulations we typically
identify $s$ with time, $s=t$, except in the truncated V--neck of Case~A3.

\paragraph{Straight, V-- and S--centerlines.}

The three centerlines used in Case~A are obtained by choosing different
offsets $\Delta(s)$ in~\eqref{eq:app_centerline_generic}:
\begin{align}
  \text{Straight:}\quad
  &\Delta_{\mathrm{lin}}(s) \doteq  0,\\[0.25em]
  \text{V--neck:}\quad
  &\Delta_{\mathrm{V}}(s)
     = -A_{\mathrm{V}}
       \bigl(1 - |2s-1|\bigr),\\[0.25em]
  \text{S--tunnel:}\quad
  &\Delta_{\mathrm{S}}(s)
     = A_{\mathrm{S}}\,\sin(2\pi s),
\end{align}
with amplitudes $A_{\mathrm{V}},A_{\mathrm{S}}>0$ chosen so that the maximal
transverse excursion $|\Delta(s)|$ is of order $1.5$ in the V-- and S--cases.
We then set
\[
  \nu^{(\mathrm{lin})}_t = \nu_{\mathrm{lin}}(s{=}t),\quad
  \nu^{(\mathrm{V})}_t   = \nu_{\mathrm{V}}(s{=}t),\quad
  \nu^{(\mathrm{S})}_t   = \nu_{\mathrm{S}}(s{=}t),
\]
unless otherwise noted.  By construction
$\nu_0=x_{\mathrm{in}}$ and $\nu_1=x_{\mathrm{out}}$ for all three templates.

For the truncated V--neck of Case~A3 we introduce a cutoff parameter
$s_{\max}\in(0,1]$ and define
\begin{equation}
  \nu^{(\mathrm{V},s_{\max})}_t
    = \nu_{\mathrm{V}}(\min\{t,s_{\max}\}),
  \label{eq:app_V_cutoff}
\end{equation}
so that the guide moves along the same V--centerline as in Case~A1 but stops
after covering a fraction $s_{\max}$ of the total arclength and remains frozen
there for $t>s_{\max}$.

\paragraph{Stiffness profiles.}

In the geometry experiment (Case~A1) and the temporal progression experiment
(Case~A3) we use a constant stiffness,
\begin{equation}
  \beta^{(\mathrm{geom})}(t)\doteq  \beta_{\mathrm{geom}}>0,\qquad
  \beta^{(\mathrm{V})}(t)\doteq  \beta_{\mathrm{V}}>0,
\end{equation}
so that $\beta_t$ is independent of geometry and of the cutoff $s_{\max}$.

In the confinement experiment (Case~A2) we fix the S--tunnel centerline
$\nu^{(\mathrm{S})}_t$ and vary only the overall stiffness level via
scale factors $\gamma>0$:
\begin{equation}
  \beta^{(\mathrm{S},\gamma)}(t)
    \doteq  \gamma\,\beta_{\mathrm{base}},
  \label{eq:app_S_beta_scaled}
\end{equation}
with $\beta_{\mathrm{base}}>0$ fixed and $\gamma$ ranging from loose to tight
confinement.  This yields a family of protocols
$\Gamma^{(\mathrm{S},\gamma)}_t=(\nu^{(\mathrm{S})}_t,\beta^{(\mathrm{S},\gamma)}_t)$
that all share the same geometry but differ in tube width.

\paragraph{PWC discretization.}

To obtain a protocol that can be fed to the GH--PID sampler we discretize
$[0,1]$ into $K$ pieces:
\[
  0=t_0<t_1<\dots<t_K=1.
\]
For each interval $[t_k,t_{k+1})$ we define its midpoint
$t_k^\star=\tfrac12(t_k+t_{k+1})$.  To ensure that both endpoints of the
continuous guide are represented exactly in the PWC protocol, we set
\begin{equation}
  \nu_0 = \nu_{t=0},\qquad
  \nu_{K-1} = \nu_{t=1},
\end{equation}
and define the interior values at midpoints,
\begin{equation}
  \nu_k = \nu_{t_k^\star},\qquad
  k=1,\dots,K-2.
  \label{eq:PWC_nu}
\end{equation}
The stiffness values are sampled at midpoints for all segments,
\begin{equation}
  \beta_k = \beta_{t_k^\star},\qquad k=0,\dots,K-1.
  \label{eq:PWC_beta}
\end{equation}
The resulting PWC protocol
$\Gamma^{(\mathrm{  PWC})}=\{(\nu_k,\beta_k)\}$ is passed to the analytic
GH--PID machinery via the PWC schedule object described in
Section~\ref{sec:formulation}.  In the navigation figures the discrete
$\nu_k$ are shown as crosses, while the continuous $\nu_t$ is shown as a dashed
curve.

\paragraph{Corridor walls for visualization.}

The corridor walls plotted in the figures are derived from the same centerline
but are used purely for visualization.  We define a width profile $w(t)>0$
(either prescribed directly or chosen to be inversely related to a reference
stiffness) and construct left/right walls by offsetting the centerline along
the unit normal $n$,
\[
  \mathrm{left}(t)  = \nu_t + w(t)\,n,\qquad
  \mathrm{right}(t) = \nu_t - w(t)\,n.
\]
In practice the walls are computed on a discrete grid in $t$ and linearly
interpolated for plotting.  These walls do \emph{not} enter the GH--PID
equations; they only serve to make the geometric interpretation of
$\Gamma_t=(\nu_t,\beta_t)$ visually transparent.

\paragraph{Simulation details.}

We simulate the GH--PID diffusion using an Euler–Maruyama discretization
of~\eqref{eq:sde} with $u_t\to u_t^\ast$ from
\eqref{eq:ustar_compact_forwardref}.  Time is discretized into $T$ equal
steps; the PWC protocol $(\nu_k,\beta_k)$ is aligned with this grid via a
mapping from the time index $n$ to the corresponding interval $[t_k,t_{k+1})$.
Unless otherwise stated we use $T\approx 10^3$--$10^4$ and Monte Carlo
ensembles of size $M=10^3$--$10^4$ particles.  Snapshots of the cloud
$\{x_t\}$ are taken at a fixed set of times, including $t=0^+$, and displayed
together with the target samples, the centerline, and the corridor walls.
Empirical means and covariance ellipses are plotted to highlight adherence to
the guide and the shape of the cloud.

\subsection{Case B: Parametric Protocols and Gradient-Based Learning}
\label{app:caseB}

Case~B equips the guidance protocol with a low–dimensional but expressive parametric structure, enabling joint learning of geometry, stiffness, and temporal allocation.  The resulting protocol family remains analytically tractable under the PWC machinery of Appendix~\ref{sec:Riccati}.

\paragraph{Spline geometry.} We work in the intrinsic corridor frame $(e,n)$ defined in Appendix~D.1.
Let $c_1=x_{\rm in}$ and $c_M=x_{\rm out}$. For intermediate control points $\{c_2,\dots,c_{M-1}\}$, collected in $\theta_\nu$, we define a cubic spline
\[
\tilde{\nu}(s;\theta_\nu)=\mathrm{Spline}(c_1,\dots,c_M)(s),\qquad s\in[0,1].
\]
The learned centerline is then $\nu(t;\theta)=\tilde{\nu}(s(t;\theta_s);\theta_\nu)$.
During execution we sample $\nu(t)$ piecewise–constantly as in~(D.9)–(D.10).

\paragraph{Learnable stiffness profile.} The stiffness is parametrized by a bounded, smooth profile
\[
\beta(t;\theta_\beta)
=
\beta_{\min} + (\beta_{\max}-\beta_{\min})
\,\sigma\!\left(a_\beta (t-t_0)\right),
\]
with $(a_\beta,t_0,\beta_{\min},\beta_{\max})\in\theta_\beta$. We enforce
\[
0<\beta_{\min}\le \beta(t;\theta_\beta)\le \beta_{\max}<\beta_{\rm stable},
\]
where $\beta_{\rm stable}$ is the empirically determined stiffness threshold from Case~A2 beyond which numerical effects dominate.

\paragraph{Time warping.} To control the temporal progression along the guide, we introduce a monotone re-timing map
\[
s(t;\theta_s)=t+\alpha\,t(1-t),
\qquad \alpha\in\theta_s,\quad |\alpha|\le \alpha_{\max},
\]
with $\alpha_{\max}$ chosen to prevent extreme compression or dilation. More expressive alternatives (two–knot and three–knot piecewise–linear warpings) may be used with the same constraints $0=s(0)<s(1)=1$ and $s'(t)\ge 0$.

\paragraph{PWC protocol construction.} Let $0=t_0<\dots<t_K=1$ be a uniform partition. Define midpoints $t_k^\star=\tfrac12(t_k+t_{k+1})$. We set
\[
\nu_k = \nu(t_k^\star;\theta),\qquad
\beta_k = \beta(t_k^\star;\theta),
\]
and assemble the PWC protocol $\Gamma^{\rm (PWC)}=\{(\nu_k,\beta_k)\}_{k=0}^{K-1}$. The GH--PID Riccati coefficients $(a^{(\pm)}_t,b_t^{(-)},c_t^{(-)})$ and linear terms $(r_t^{(-)},s_t^{(-)},s_t^{(+)})$ are then computed exactly using the update rules of Appendix~\ref{sec:Riccati}.

\paragraph{Differentiability.} All components—spline geometry, stiffness, warping, PWC sampling, Green function updates, and drift evaluation—are differentiable almost everywhere with respect to~$\theta$. The only non–smooth events are the PWC jumps at grid points $t_k$, whose effect on
gradients is handled automatically by modern autodiff systems when implemented via coordinate–wise operations.

\paragraph{Objective and optimization.} We optimize the multi–term objective~(\ref{eq:caseBobjective}) from Section~\ref{subsec:caseB}, combining adherence, control effort, barrier–based landscape penalties, curvature regularization, and warp regularization. Gradients are estimated via Monte–Carlo GH–PID trajectories in PyTorch, and $\theta$ is updated using Adam with box constraints enforcing bounds on $(\beta_{\min},\beta_{\max})$ and $|\alpha|\le\alpha_{\max}$. The resulting learned protocol is evaluated using the diagnostics of Section~3 and compared against hand–crafted protocols in Section~\ref{subsec:caseA}.

\subsection{Case C: Product-of-Experts Targets and Multi-Task Optimization}
\label{app:caseC}

In Case~C we consider $M$ navigation tasks that share the same class of
protocols but differ in their terminal GMMs and path--dependent costs.  For
task $m$ let
\[
  p^{(m)}_{\mathrm{tar}}(x)
   = \sum_{j=1}^{J_m} \pi^{(m)}_j\,
       \mathcal{N}(x\mid\mu^{(m)}_j,\Sigma^{(m)}_j)
\]
be its terminal law.  We form a fused terminal density by a
product--of--experts construction,
\begin{equation}
  p_{\mathrm{tar}}(x)
  \propto \prod_{m=1}^M
      \bigl(p^{(m)}_{\mathrm{tar}}(x)\bigr)^{\alpha_m},
  \label{eq:poe_target}
\end{equation}
with nonnegative weights $\alpha_m$ that tune the relative influence of each
task.  Because products of Gaussians are Gaussian, the fused density remains a
GMM with updated means, covariances, and weights that can be computed in
closed form by standard Gaussian product identities.  This keeps the GH--PID
expressions for $\hat{y}(t;x)$ and $u_t^\ast(x)$ fully analytic.

The multi--task protocol optimization problem is formulated as
\[
   \mathcal{J}_{\mathrm{multi}}(\Gamma)
     = \sum_{m=1}^M \lambda_m
         \int_0^1 \mathbb{E}\big[U^{(m)}_t(x_t)\mid \Gamma\big]\,dt,
\]
with $\lambda_m\ge 0$ representing task priorities and $U^{(m)}_t(x)$ the
task--specific landscape energies derived, for example, from risk maps or
region--of--interest fields.  In practice we reuse the parametric family
$\Gamma(t;\theta)$ from Case~B and optimize the same parameter vector $\theta$
against the composite cost $\mathcal{J}_{\mathrm{multi}}(\Gamma(\theta))$,
again using Monte Carlo estimates and automatic differentiation to obtain
gradients.

This construction shows that a single analytic GH--PID backbone can support
multi--task navigation: several terminal objectives and cost landscapes are
fused into a single SOT problem with a product--of--experts hard constraint and
a weighted combination of soft path costs.

\end{document}